\documentclass[Afour,sagev,times]{sagej}

\usepackage{moreverb,url}

\usepackage{nameref}

\newcommand\BibTeX{{\rmfamily B\kern-.05em \textsc{i\kern-.025em b}\kern-.08em
T\kern-.1667em\lower.7ex\hbox{E}\kern-.125emX}}

\def\volumeyear{2024}

\usepackage[caption=false,font=normalsize,labelfont=sf,textfont=sf]{subfig}
\usepackage{url}
\usepackage{graphicx}
\usepackage{booktabs}
\usepackage{array}
\usepackage{tabularx}
\usepackage{bm}
\usepackage{lscape}
\usepackage{comment}
\usepackage{enumitem}%
\graphicspath{{figures/}}
\usepackage{amsmath}
\usepackage{lipsum}
\usepackage{mathtools}
\usepackage{float}

\newcolumntype{s}{>{\hsize=1.5cm}X}
\usepackage{nicefrac}
\usepackage{cite}
\usepackage{graphicx}
\usepackage{caption}

\usepackage{array}
\newcommand{\PreserveBackslash}[1]{\let\temp=\\#1\let\\=\temp}
\newcolumntype{C}[1]{>{\PreserveBackslash\centering}p{#1}}
\newcolumntype{R}[1]{>{\PreserveBackslash\raggedleft}p{#1}}
\newcolumntype{L}[1]{>{\PreserveBackslash\raggedright}p{#1}}
\newcolumntype{P}[1]{>{\centering\arraybackslash}p{#1}}

\usepackage{booktabs} %
\usepackage{siunitx} %
\usepackage{tabularx} %
\usepackage{makecell} %
\usepackage{booktabs} %
\usepackage{amsmath}
\usepackage{graphicx}
\usepackage[table,xcdraw]{xcolor}
\usepackage{booktabs}
\usepackage{array}
\usepackage[colorlinks=false, hidelinks]{hyperref}
\usepackage{multirow}
\usepackage{tabularx}

\makeatletter
\def\@maketitle{%
\if@Royal
\vspace*{-20pt}
\fi
\if@Crown
\vspace*{-20pt}
\fi
\vspace*{-34pt}%
\null%
\begin{center}
\if@PCfour
\begin{rm}
\else
\begin{sf}
\fi
\begin{minipage}[t]{\textwidth-57.625mm}
  \vskip 12.5pt%
    {\raggedright\titlesize\textbf{\@title} \par}%
    \vskip 1.5em%
    \vskip 12.5mm%
    \end{minipage}\hspace{15mm}\begin{minipage}[t]{42.625mm}
\raggedright\scriptsize
\textbf{Accepted manuscript}\\[2pt]
Published in \textit{\journalname}\\[2pt]
\copyright\ The Author(s) 2025\\[2pt]
DOI: \href{https://doi.org/10.1177/15485129241309675}{10.1177/15485129241309675}
\end{minipage}
{\par\large%
\if@Royal
      \vspace*{6mm}
      \fi
      \if@Crown
      \vspace*{6mm}
      \fi%
      \lineskip .5em%
      {\raggedright\textbf{\@author}
      \par}}
     \vskip 40pt%
    {\noindent\usebox\absbox\par}
    {\vspace{20pt}%
      {\noindent\normalsize\@keywords}\par}
      \if@PCfour
      \end{rm}
      \else
      \end{sf}
      \fi
      \end{center}
      \if@Royal
      \vspace*{-4.5mm}
      \fi
      \if@Crown
      \vspace*{-4.5mm}
      \fi
      \vspace{22pt}
        \par%
  }
\def\ps@title{%
\def\@oddhead{\parbox{\textwidth}{\mbox{}\\[-1pt]%
\noindent\rule{\textwidth}{0.5pt}%
}}%
\let\@evenhead\@oddhead
\let\@oddfoot\@empty
\let\@evenfoot\@empty}
\def\ps@sagepage{%
\let\@mkboth\@gobbletwo
\def\@evenhead{\parbox{\textwidth}{%
\normalsize\sagesf\thepage\hfill\itshape Accepted manuscript\\[-6pt]
\noindent\rule{\textwidth}{0.25pt}}}
\def\@oddhead{\parbox{\textwidth}{%
\normalsize\sagesf{\itshape{\leftmark}}\hfill\thepage\\[-6pt]
\noindent\rule{\textwidth}{0.25pt}}}
\let\@evenfoot\@empty
\let\@oddfoot\@empty
}
\makeatother

\begin{document}

\runninghead{Dantas et al.}

\title{PoKER: A Probability of Kill Estimation Rate Model for Air-to-Air Missiles Using Machine Learning on Stochastic Targets}

\author{Joao P. A. Dantas\affilnum{1}, Andre N. Costa\affilnum{1}, Diego Geraldo\affilnum{1}, Marcos R. O. A. Maximo\affilnum{2}, Takashi Yoneyama\affilnum{3}}

\affiliation{
\affilnum{1}Decision Support Systems Subdivision, Institute for Advanced Studies Brazil\\
\affilnum{2}Computer Science Division, Aeronautics Institute of Technology, Brazil
\affilnum{3}Electronic Engineering Division, Aeronautics Institute of Technology, Brazil
}

\corrauth{Joao P. A. Dantas, Institute for Advanced Studies, Decision Support Systems Subdivision, Trevo Coronel Aviador Jose Alberto Albano do Amarante, 1, Sao Jose dos Campos, Sao Paulo, 12.228-001, Brazil}

\email{jpdantas@ita.br}

\begin{abstract}
This work introduces PoKER, a novel probabilistic model engineered to optimize missile launch effectiveness in air-to-air scenarios, specifically within Beyond Visual Range (BVR) air combat. Unlike conventional Weapon Engagement Zone (WEZ) models that delineate zones based on static distances such as maximum, minimum, and no-escape ranges, PoKER applies machine learning to predict kill probabilities more accurately by integrating the stochastic behaviors of targets and missile miss distances. This model dynamically evaluates target behavior, greatly expanding the predictive capabilities of engagement analysis. By factoring in elements such as warhead lethality, target and shooter orientations, and the specific conditions of engagement, PoKER provides important insights into engagement dynamics and quantifies success probabilities. Consequently, it can potentially be an important tool for BVR air combat pilots, improving operational decision-making within this specialized combat domain.
\end{abstract}

\keywords{Air Combat, Decision Support Systems, Machine Learning, Military Simulation, Weapon Engagement Zone}

\maketitle

\footnotetext[0]{\sagesf\textbf{Accepted manuscript.} This is the accepted manuscript of an article published in \textit{The Journal of Defense Modeling and Simulation: Applications, Methodology, Technology}. Copyright \copyright\ 2025 The Author(s). DOI: \href{https://doi.org/10.1177/15485129241309675}{10.1177/15485129241309675}. This version has not undergone the publisher's copyediting and typesetting; please cite the published version.}

\section{Introduction} \label{introduction}

Missile systems hold a critical position in contemporary military operations~\citep{li2020simulation}, providing an effective means to engage aerial targets over extensive distances, particularly significant in the context of Beyond Visual Range (BVR) air combat. BVR air combat refers to aerial engagements that occur at distances beyond the visual sight of the pilots, typically over tens of kilometers. In these scenarios, aircraft rely on advanced radar, sensors, and long-range missiles to detect, track, and engage opponents without the need for visual identification~\citep{dantas2021engagement}. The evolving landscape of BVR air combat, driven by recent advancements in detection and missile technology, underscores the necessity for sophisticated Weapon Engagement Zone (WEZ) models~\citep{li2022multi}. These models, essential for evaluating the operational efficacy of air-to-air missiles, bring about complex computations of maximum and minimum engagement ranges alongside the identification of No Escape Zones (NEZ)~\citep{dantas2021weapon}. Notably, these computations are usually dynamically executed in real-time during flight, factoring in the orientations of both the target and the shooter~\citep{herrmann1996air}.

WEZs are not simply geometric boundaries determined by orientations and positions; they also take into account the dynamics and maneuverability of both the target and shooter, along with their weapons systems~\citep{dantas2021engagement}. The insights derived from WEZ calculations may be relayed to pilots through heads-up or heads-down displays~\citep{yoon2010new}, which may empower them to make more informed decisions regarding missile launches with increased precision and confidence~\citep{ling2019cognitive}.

\begin{figure}[htbp]
\centering
\includegraphics[width=0.485\textwidth, trim={2.3cm 1.1cm 2.3cm 1.1cm}, clip]{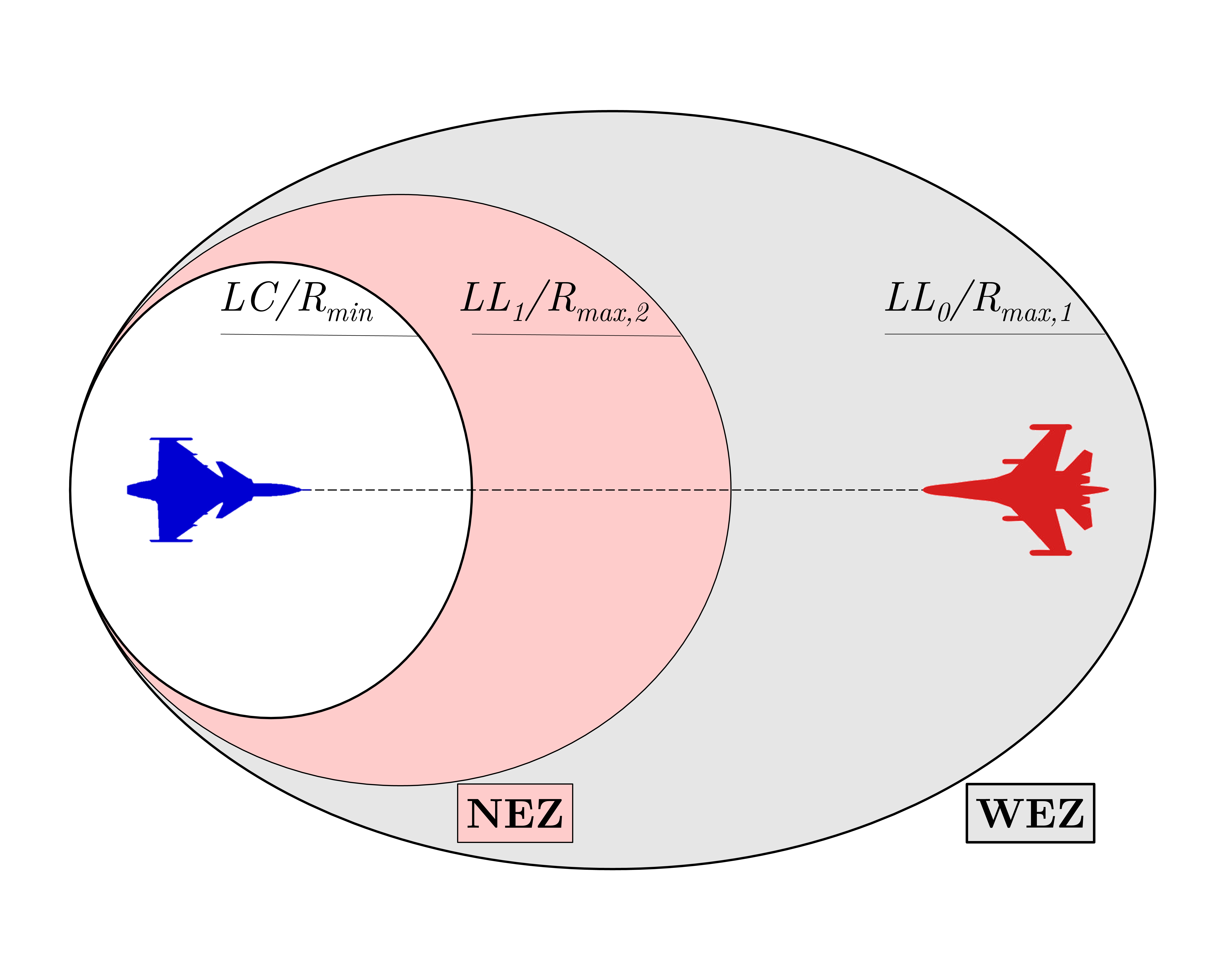}
\caption{Diagram illustrating aerial combat WEZ, detailing ranges for missile engagement from maximum kinematic reach to minimum arming distance, including the airspace where the target has limited evasion options.}
\label{fig:weznez}
\end{figure}

The definition of WEZ can incorporate concepts such as long and short limits or maximum and minimum ranges, whose interpretations might slightly vary between countries~\citep{van2019optimizingy}. To exemplify, we present two common nomenclatures, as depicted in Figure~\ref{fig:weznez}, to evidence their nuances, mostly with respect to enemy maneuverability. Some armed forces use Long Limit 0 ($LL_0$) to delineate the furthest kinematic reach of a missile, where it uses up all its energy just to activate its proximity fuse upon reaching the target. Similarly, Maximum Range ($R_{max,1}$) indicates the maximum distance a missile can effectively engage a passive target, with its success probability decreasing if the target maneuvers to increase flight distance. Long Limit 1 ($LL_1$) specifies the range within which a missile can effectively counteract a target's ultimate evasion efforts with maximum performance. Alternatively, $R_{max,2}$ introduces the beginning of the NEZ, which represents a range within which the missile is expected to hit, barring specific defensive actions by the target, though escape is still possible through engine maneuvers or missile countermeasures. Lastly, the Short Limit ($LC$)/$R_{min}$ marks the closest distance at which a missile's fuze can arm.

Even though these concepts provide a general idea of where to engage a target, depending on rules of engagement and shot philosophies~\citep{dantas2022supervised, kuroswiski2023beyond}, these regions may still be quite extreme, not accounting for what happens in between them, which makes it difficult for pilots to choose the best moment to launch a missile in order to maximize the chances of success. This is aggravated by the fact that the opposing aircraft may maneuver unpredictably. Amid these complex considerations, it becomes essential to conduct detailed simulations to improve operational readiness and tactical accuracy, analyzing missile efficacy in diverse engagement scenarios, which include different adversarial reactions to missile threats. In this research, we introduce a novel WEZ model tailored to improve the assessment of missile performance after launch, specifically in the case of an Active Radar Homing (ARH) weapon system utilized in BVR air combat. Leveraging insights from prior work, we present an R-based simulator based on a 5 Degrees of Freedom (5DOF) model~\citep{handbook1995missile}.

ARH, or Active Radar Homing, refers to a guidance method used in missile systems, where the missile itself is equipped with a radar seeker that actively emits radar signals to detect and track its target independently. This type of homing provides the missile with greater autonomy after launch, reducing the need for continuous guidance from the launching platform and enabling effective engagement of targets in BVR air combat scenarios.

Our simulator encapsulates the complexities of an ARH missile, including its radar seeker functionality and autonomous target-tracking capability, enabling rigorous analysis and prediction of its performance across various operational parameters. By exploring the nuances of missile-target interactions with an active radar seeker, our simulator provides valuable insights into BVR air combat dynamics, aiding in the formulation of strategic and tactical decisions.

Furthermore, we explore the domain of warhead lethality, offering insights into the calculation of the proximity-fuzed warhead's conditional probability of target destruction, a critical component in assessing missile effectiveness. Additionally, we focus on generating and preprocessing simulation data, proposing a supervised machine learning methodology to develop regression models capable of predicting missile kill probabilities. We carefully evaluate various algorithms' efficacy and computational efficiency, illuminating their suitability for real-world applications.

The main contribution of this work is as follows:

\begin{itemize}
    \item \textbf{Development of the Probability of Kill Estimation Rate (PoKER) model}: Designed to provide enhanced situational awareness in dynamic air-to-air combat scenarios, particularly in BVR air combat settings.
\end{itemize}

The specific contributions that support the development and application of the PoKER model are:

\begin{itemize}
    \item \textbf{Complementation of WEZ with kill probability estimation}: Addressing the gap in traditional weapon engagement analysis by providing an estimation of the probability of kill when launching air-to-air missiles, incorporating real-time conditions of both the launcher and the target for a more comprehensive understanding of engagement success.
    \item \textbf{Modeling target decisions}: Integration of potential adversary maneuvers using high-fidelity models of aircraft and missiles, allowing for flexibility rather than relying on fixed target profiles and predefined launcher strategies.
    \item \textbf{Simulation and machine learning integration}: Utilization of well-designed simulation tools and machine learning to deepen the understanding of missile engagement dynamics.
    \item \textbf{Warhead lethality model}: Inclusion of a warhead lethality model that estimates the probability of kill ($P_{\text{kill}}$) based on missile miss distance and a dynamic model of the opponent’s behavior, substantially improving tactical decision-making and enhancing operational capabilities in BVR air combat.
\end{itemize}

In summary, the primary focus of the PoKER model is to enhance situational awareness in dynamic BVR engagements, marking an initial step towards the development of probabilistic models for weapon engagement analysis. This work offers a novel methodological contribution, advancing beyond traditional WEZ analysis by incorporating real-time probabilistic assessments and accounting for the dynamic interactions between the launcher and the target. Furthermore, the insights gained from this model could be leveraged in the future to support decision-making processes, enabling more informed tactical decisions in complex air combat scenarios.

The remainder of this document is organized as follows. The~\nameref{related-work} section provides an overview of the literature pertinent to our study and introduces critical terminology. The~\nameref{methodology} section delineates our approach, outlining each of its steps and the employed tools and techniques. The~\nameref{results} section showcases and discusses the findings derived from the application of our methodology. The~\nameref{conclusion} section summarizes our approach and its outcomes while also highlighting potential avenues for future research.
\section{Related Work}\label{related-work}

The success of BVR air combat heavily relies on the seamless integration of offensive and defensive tactics supported by accurate and timely information. Consequently, numerous research efforts have focused on enhancing tactical decision-making through advanced air combat simulations, aiming to improve the effectiveness of engagements in dynamic scenarios~\citep{dantas2023autonomous, dantas2022machine, reinisch2022behaviour, ummah2019simple, kang2019beyond}. These studies explore various aspects of BVR air combat, including weapon engagement, evasion strategies, and cooperative tactics, all of which contribute to a more comprehensive understanding of air combat dynamics.

\subsection{Engagement and Escape Zone Strategies}

The current study models predominantly integrate missile information to define Dynamic Escape Zones (DEZ), which refers to the region where the aircraft's survivability can be guaranteed through kinetic evasion maneuvers~\citep{yagci2022evasive}, and WEZ, which are also referred to as Dynamic Launch Zones~\citep{katukuri2023target}, Missile Attack Zones~\citep{qian2022missile}, or Launch Acceptability Regions~\citep{ozdemir2021ai}. The primary aim of WEZ models is to aid in decision-making processes concerning weapon launch and guidance support -- through uplink updates -- by outlining the boundaries and operational effectiveness of air-to-air missiles during target engagement. Two main methods are traditionally used to calculate WEZ for various missile configurations. The first method conducts ongoing flyout simulations of the missile model during the engagement period. The second, more widely used method performs simulated flyouts in an offline setting, creating a lookup table that incorporates the pre-simulated data. Upon employing the WEZ model, interpolation is applied to this data to produce outcomes that are not directly available in the table~\citep{birkmire2011weapon}.

Several techniques have been developed for both collecting the predefined data and for its subsequent interpolation. However, notable advancements have also been made in fields such as post-launch WEZ analysis, cooperative modeling of WEZ, and algorithms for WEZ estimation~\citep{hui2015dynamic, shi2018modeling, yue2020deep, dantas2021weapon}. These studies tailor WEZ concepts to specific combat scenarios and maneuver types, often integrating them with three-degree-of-freedom (3-DOF) or six-degree-of-freedom (6-DOF) models~\citep{venkatesan2015new}. Solution methods for WEZ calculations range from offline simulation algorithms to polynomial, interpolation, and neural network fitting algorithms~\citep{dantas2023real}.

On the other hand, the DEZ definition aims to encompass essential elements of tactical decision-making from the perspective of missile avoidance. Techniques such as differential game theory and principles like Minimum Evasive Range~\citep{kung2015study} or NEZ~\citep{scukins2023monte} play a central role in differentiating between aggressive and defensive tactics, as well as in identifying the best timing for evasive actions~\citep{alkaher2015dynamic, ryu2018no}. This can be done either by analyzing missile miss distances in one-on-one engagements or by considering multiple incoming threats~\citep{scukins2024deep}.

\subsection{Cooperative Engagement Models}

While most of the earlier works focused on duel engagements, there has been a shift towards modeling cooperative engagement among multiple aircraft, particularly in scenarios involving Unmanned Combat Aerial Vehicles (UCAVs)~\citep{li2020simulation}. Cooperative engagement models extend WEZ and DEZ strategies by introducing coordination between multiple aircraft to enhance engagement effectiveness and survivability. These models aim to optimize WEZ and NEZ calculations using cooperative strategies, leveraging algorithmic approaches to increase effectiveness in group engagements~\citep{li2022multi}. Furthermore, optimization methodologies for attack positioning and cooperative tactics have been proposed, incorporating offensive and defensive factors for assessing the best maneuvers~\citep{yang2014best, gao2022real, gao2023design}.

\subsection{Probabilistic Modeling in BVR Air Combat}

In recent years, there has been growing interest in incorporating probabilistic elements into BVR combat modeling, moving beyond traditional deterministic approaches. Some works have used Bayesian Networks to create probabilistic graphical models representing variables and their conditional dependencies through a directed acyclic graph~\citep{sun2020modeling}. BVR combat can be divided into four phases: position occupy maneuver, launch maneuver, midcourse guidance, and terminal guidance. Utilizing Dynamic Bayesian Networks (DBN), these studies encapsulate the probabilistic relationships and uncertainties in each phase, predicting the likelihood of a successful missile hit by considering aircraft positions, velocities, maneuvers, missile specifications, and environmental conditions. Integrating expert knowledge and data from air combat simulations creates a temporal representation of the combat scenario. Through DBN analysis, key determinants of missile success probability are identified, and the Dynamic Attack Zone (DAZ) is delineated -- representing the region where a missile launch is likely to result in a hit. This probabilistic framework facilitates simulation and tactical analysis, providing insights for optimizing missile deployment and supporting decision-making under uncertainty in BVR engagements.

\subsection{Limitations in Existing Work}

While the existing literature explores various dimensions of WEZ determination, a noticeable gap exists regarding incorporating the target's maneuvering capabilities into WEZ creation. Even some works that include probabilities through Bayesian Networks do not show clear variability in the target maneuvering, as the target makes a turn as soon as it detects the threat of the incoming missile~\citep{sun2020modeling}. Our approach distinguishes itself by incorporating the concept of maneuver probability into the process of determining the WEZ, a feature absent in the studies we reviewed. This innovative method extends beyond only considering the initial dynamics and physical attributes of both the missile and the aircraft. It proactively forecasts the adversary's potential maneuvers, transforming the WEZ from a simple geometric entity into a dynamic probability distribution that reflects the likelihood of the missile's successful engagement with the target. Our model furthermore enhances this by incorporating the target's behavior in a probabilistic manner and providing a probability of success through studying warhead lethality.
\section{Methodology}\label{methodology}

This section examines the methodology encompassing the missile launch simulator, the acquisition of simulation data, and the preprocessing techniques applied to ensure high-quality datasets. We thoroughly describe the input variables and the steps taken to analyze and undersample the data. We then elucidate the methods used to calculate warhead lethality, which is integral to determining the missile's probability of kill based on its miss distance. The development, training, and evaluation of supervised machine learning models are discussed in depth, concluding with an assessment of their performance. Finally, we explore the estimation of turn degree and delay, key factors in computing the probability of hitting a target with the missile.

\subsection{Missile Launch Simulator}

In this research, we have developed an enhanced version of the missile launch simulator, primarily based on our previous work~\citep{dantas2021weapon}. Our simulator, implemented in R, models the behavior of a Fox 3 missile, following a 5DOF framework as outlined in the Missile Handbook~\citep{handbook1995missile}. This type of missile, as defined in military terminology~\citep{brevity2020}, is an ARH (Active Radar Homing) missile, meaning it is equipped with an autonomous seeker capable of tracking a target after activation at a specific range~\citep{dantas2021weapon}. Through this simulator, we are able to analyze and predict missile behavior and performance under a variety of operational conditions, providing valuable insights into missile-target interactions in BVR air combat. Due to confidentiality and sovereignty concerns, the missile model is under restricted access.

\subsubsection{Guidance and Navigation} The model's core feature is its guidance system, designed to emulate proportional navigation. This ensures the missile maintains an optimal trajectory relative to the moving target, adjusting its course to align precisely according to its guidance law. Furthermore, the simulator incorporates the possibility of a loft maneuver - a sharp, upward trajectory immediately after launch. This maneuver is critical in extending the missile's range and enhancing its probability of intercepting distant targets~\citep{dantas2021weapon}.

\subsubsection{Target Interaction} 
\begin{figure}[ht]
\centering
\includegraphics[width=\linewidth, trim={0.1cm 0.0cm 0.1cm 0.1cm}, clip]{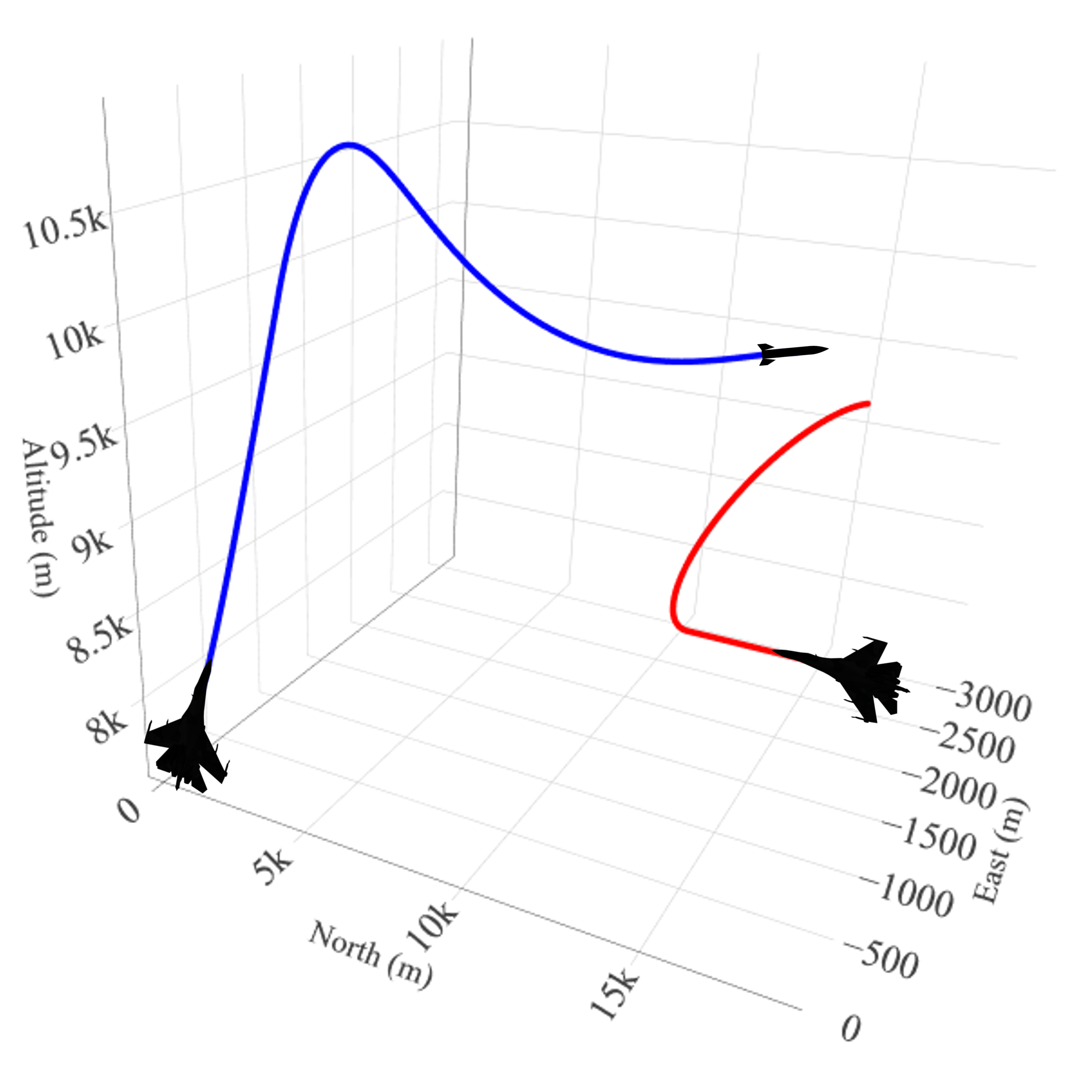}
\caption{Missile engaging a target performing a +5 G maneuver, illustrating the missile's adaptive trajectory in response to the target's high-G evasive actions.} \label{fig:target}
\end{figure}

\begin{figure}[ht]
\centering
\includegraphics[width=0.485\textwidth, trim={0cm 0cm 10.5cm 0cm}, clip]{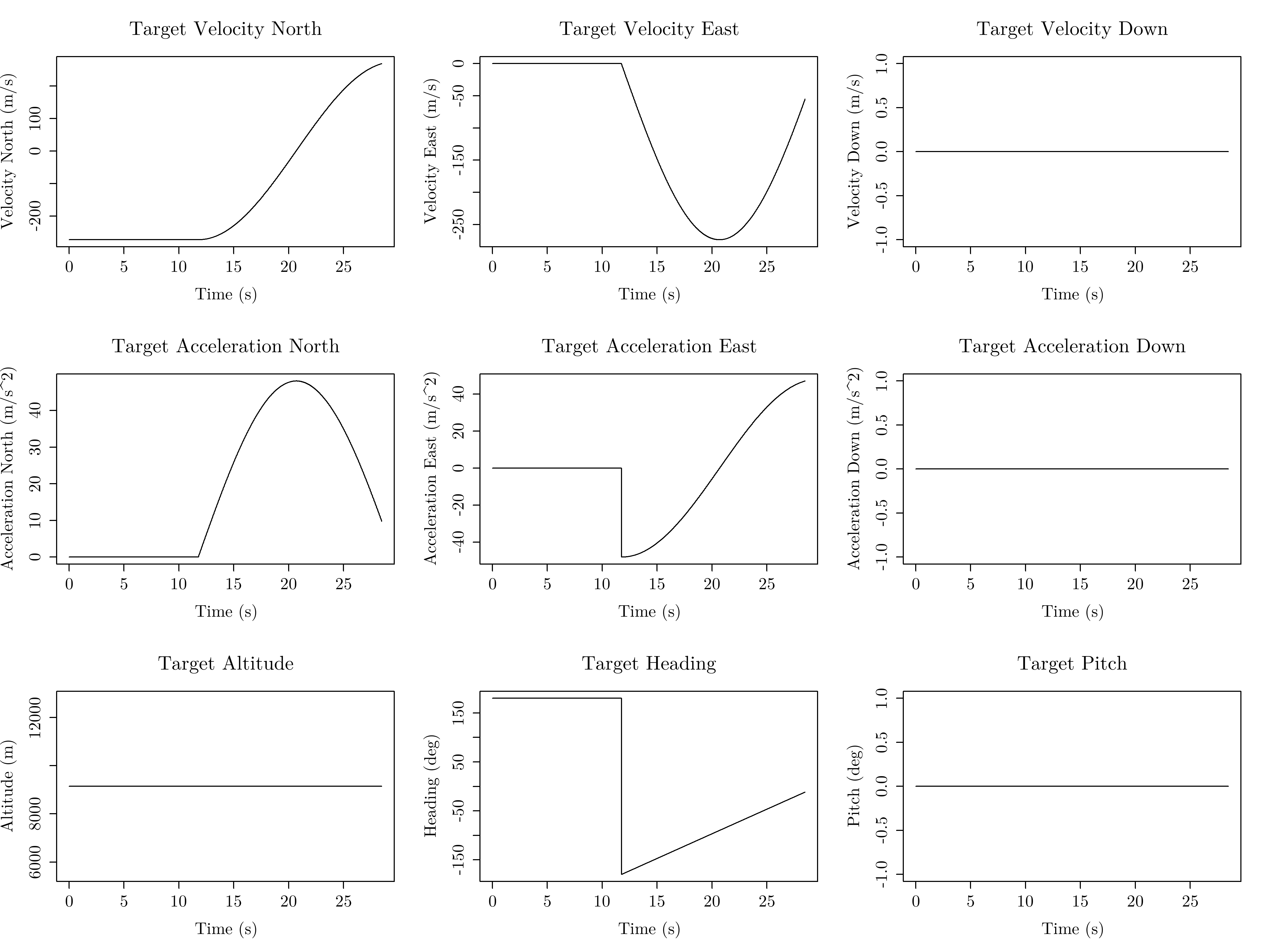}
\caption{Metrics from the simulation include target velocity (North, East), acceleration (North, East), altitude, and heading over time. Note that in this case, the target maneuver was considered to be leveled, and the target kept a constant altitude while evading the missile, resulting in a flat altitude profile. There is also a delay time in the pilot's reaction, causing all of the charts to present a flat profile for the first few seconds.}\label{fig: target_data}
\end{figure}

The simulation considers various target behaviors, including both stationary and maneuvering targets. One key aspect of the model is its ability to simulate high-performance maneuvers of the target, such as the +5 G maneuver depicted in Figure~\ref{fig:target}. The timing of these maneuvers can be varied, allowing for a comprehensive analysis of missile performance under different engagement scenarios. The data in Figure~\ref{fig: target_data} offers insights into the target's behavior, illustrating sample metrics from the simulation, including velocity, acceleration, altitude, heading, and pitch, and showing how these maneuvers impact the engagement.

\subsubsection{Missile Trajectory Metrics}

The simulation generates a comprehensive analysis of missile trajectory metrics. One notable aspect is the missile's mass, which shows a consistent decrease almost linearly during the boost phase. This pattern aligns with the expected behavior of a dual-thrust rocket motor, a critical feature in missile design~\citep{handbook1995missile, aamir2012robust}. Additionally, the simulation reveals significant variations in the missile's angle of attack and pitch angle. These variations are especially pronounced during complex maneuvers such as the loft maneuver~\citep{ma2015high}. The data depicted in Figure~\ref{fig: missile_data} highlights these dynamics, providing a comprehensive view of the missile's performance metrics throughout the simulation.

\subsubsection{Flight Dynamics}
\begin{figure*}[ht]
\centering
\includegraphics[width=\textwidth, clip]{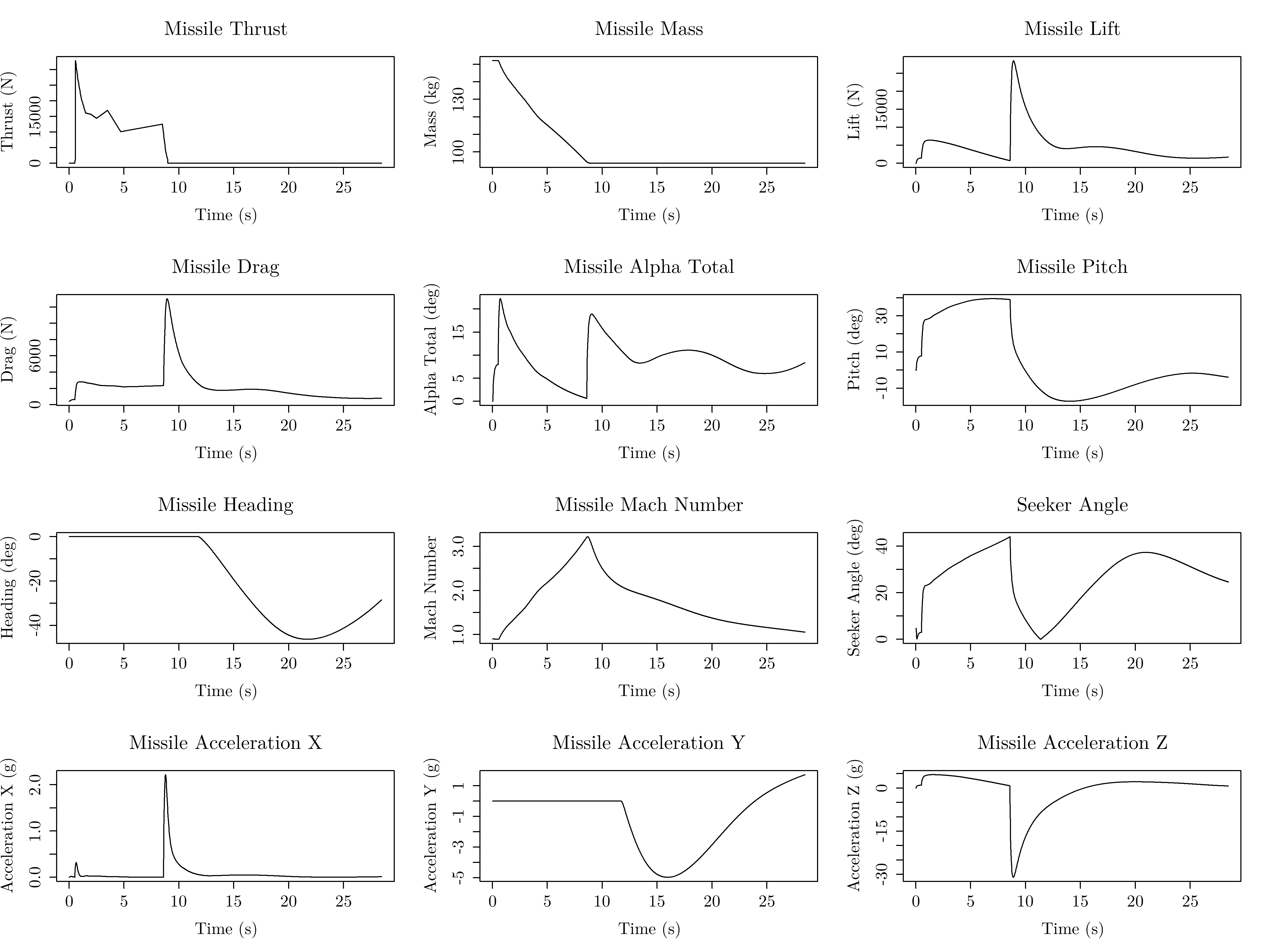}
\caption{Simulation metrics for the missile, showing thrust, mass, lift, drag, total alpha, pitch, heading, Mach number, seeker angle, and acceleration (X, Y, Z) over time.}\label{fig: missile_data}
\end{figure*}

In the domain of flight dynamics, the simulation includes detailed heading adjustments. These adjustments are fine-tuned to simulate responses to potential evasion tactics employed by a target, thereby evaluating the missile's adaptability in dynamic combat environments. The model also tracks accelerations along the East and North axes, adhering to the North-East-Down (NED) coordinate system. This aspect of the simulation is instrumental in assessing the missile’s response to both loft maneuvers and evasive actions by the target. Another key aspect studied is the velocity profile of the missile. This profile demonstrates a characteristic increase during the boost phase followed by a gradual decrease during the sustain phase~\citep{kumar2010studies}. Figure~\ref{fig: missile_data} also illustrates these aspects, providing additional insights into the missile's adaptability and performance in various dynamic scenarios.

\subsubsection{Seeker Behavior}
The final aspect covered by the simulation is the missile's seeker behavior. The seeker angle is particularly important, as it directly correlates with the missile's guidance system performance. This angle reflects the adjustments made by the missile’s seeker in following a proportional navigation path. Analyzing deviations, especially from the off-boresight angle, provides critical insights into the missile’s effectiveness, especially in scenarios where the target employs advanced evasive maneuvers~\citep{park2016range}.

\subsubsection{Visualization and Debugging Platform}

The Aerospace Simulation Environment, or \emph{Ambiente de Simulação Aeroespacial (ASA)} in Portuguese, serves as a critical resource for the Brazilian Air Force, facilitating enhanced missile launch visualization and debugging processes~\citep{dantas2022asa}. This comprehensive C++ simulation framework, tailor-made for Brazilian military applications, supports detailed examinations and visualizations of various military operations, including missile launches. More than a tool for basic training, ASA is key for executing complex, integrated combat scenarios and is essential in testing and validating tactical strategies. It provides a platform where analysts and engineers can perform comprehensive assessments and refine aerospace mechanisms and missions.

The missile model used in this research follows the same structural framework as the model used in ASA. This demonstrates the level of fidelity aimed for in this work, ensuring consistency with established simulation models in ASA. The ASA platform plays an essential role in validating and debugging this missile model, as it incorporates other elements and behaviors into the scenario, providing a more comprehensive environment for testing and refinement.

The advanced functionalities of ASA lead to improved accuracy in mission analysis and bolster reliability in operational execution~\citep{dantas2023asasimaas}. Looking forward, the insights gained from WEZ analyses in this research could be integrated into more complex simulations within ASA, where agents could make decisions based on a variety of available models, thus enhancing the overall operational readiness and tactical efficiency of the aerospace forces.

\subsection{Simulation Data} \label{sub:simulation-data}

Initially, to generate the simulation data, we created simulator input files using Latin Hypercube Sampling (LHS), an efficient statistical method for designing experiments. This method divides the space into a prespecified number of sections and randomly samples one point from each~\citep{mckay2000comparison, helton2003latin}. Unlike factorial designs, LHS offers a superior alternative for populating the sample space more efficiently than purely random methods~\citep{husslage2011space, dantas2022supervised}. We employed LHS using the AsaPy Library~\citep{dantas2024asapy}, a custom-made Python library associated with ASA, specifically designed to optimize the analysis of simulation data.

\begin{figure}[ht]
\centering
\includegraphics[width=0.485\textwidth, trim={0.5cm 0.6cm 0.5cm 0.75cm}, clip]{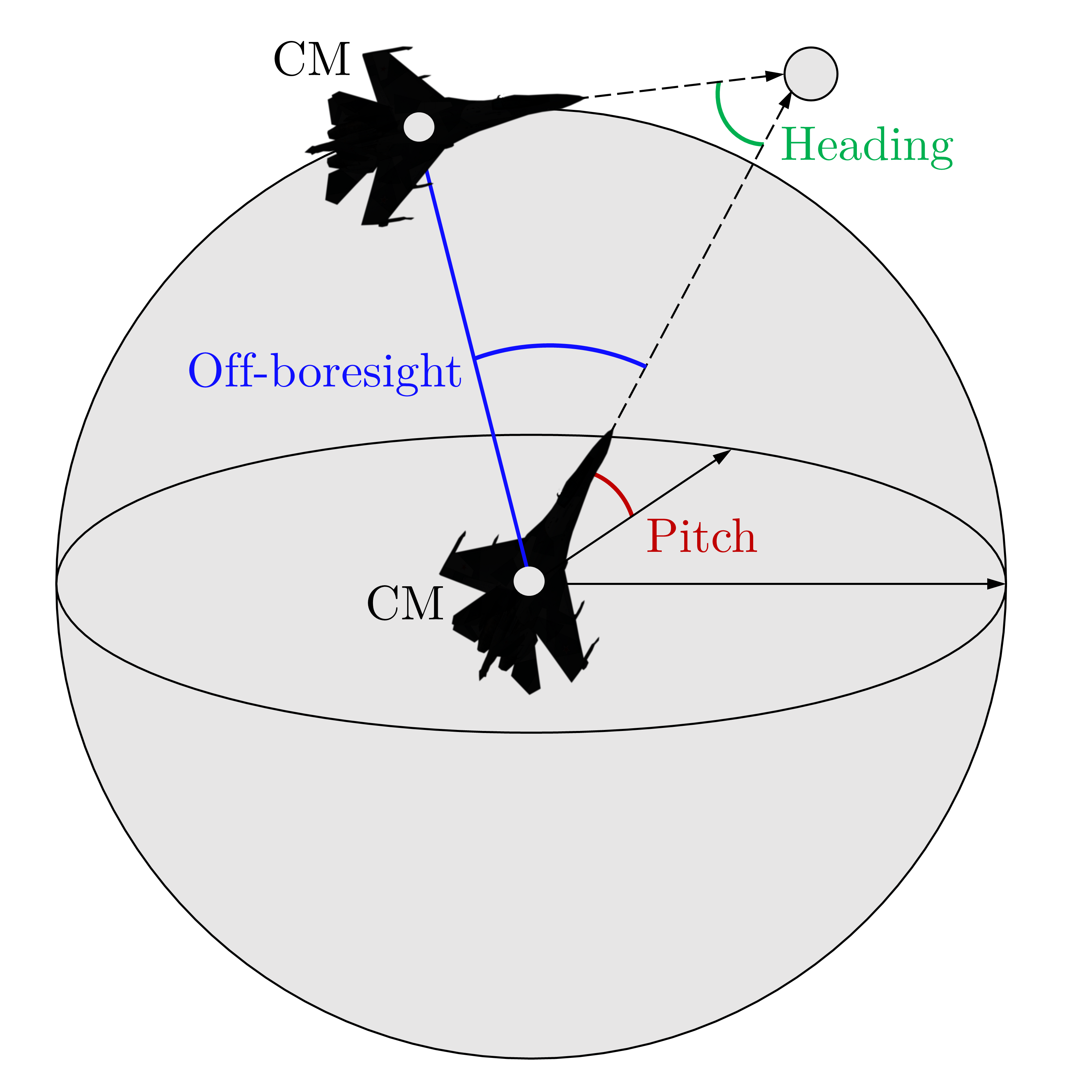}
\caption{Diagram illustrating the heading, off-boresight, and pitch angles between two aircraft in aerial maneuvering with respect to their centers of mass (CM).}\label{fig: angles}
\end{figure}

Both the shooter's altitude (\texttt{alt\_sht}) and velocity (\texttt{vel\_sht}) significantly impact the energy transferred to the launched missile~\citep{herrmann1996air}. The launch altitude (\texttt{alt\_sht}) influences the missile's drag, thus affecting flight performance~\citep{dantas2021weapon}. The shooter's pitch (\texttt{pit\_sht}), which is defined with respect to the North-East plane, is crucial for determining the missile's initial angle post-launch, aiding in the loft maneuver, which is the missile's initial maneuver after launch. The target's velocity (\texttt{vel\_tgt}) is significant, as it determines the defensibility of the launched missile. Two aspects are fundamental in terms of the angles between the shooter and the target: heading and off-boresight (Figure~\ref{fig: angles}). The target's heading (\texttt{hdg\_tgt}) indicates the direction taken to evade the missile with respect to the North, while the off-boresight angle (\texttt{rgt\_tgt}), defined as the angle between the shooter's nose and the target, along with the target's heading, reveals whether the shooter aircraft is approaching the target and thus the missile~\citep{vergez1982optimal, vergez1998tactical}. Additionally, the distance between the shooter and the target (\texttt{dist}) is critical as it determines the missile's flight time and significantly influences the hit probability. The target's response time (\texttt{delay}) represents the time before initiating a defensive maneuver. A longer response time may force the target to execute more aggressive maneuvers, which are characterized by a higher turn degree (\texttt{turn\_dg}) and greater acceleration (\texttt{load\_factor}). All measurements are taken with respect to the center of mass of either the aircraft or the missile.

Table~\ref{tb1} illustrates the range for each parameter considered during the generation of samples. These parameters were established based on the expertise of military fighter pilots, serving as Subject Matter Experts (SMEs) in military tactics, who identified critical values pertinent to the scenario. This approach aligns with methodologies previously reported in similar studies focusing on input variables within this domain~\citep{dantas2021weapon, birkmire2011weapon}.

\begin{table}[ht]
\scriptsize
\centering
\caption{Input simulation data with the respective intervals.}\label{tb1}
\begin{tabular}{|c|c|c|c|}
\hline
\textbf{Parameter}& \textbf{Min} & \textbf{Max} & \textbf{Unit} \\
\hline
Shooter altitude (\texttt{alt\_sht})        &  1,000 & 45,000   & ft    \\
Shooter velocity (\texttt{vel\_sht})        &  300  & 700             & kt   \\
Shooter pitch (\texttt{pit\_sht})           &  -30  & 30                 & deg \\
Target altitude (\texttt{alt\_tgt})         &  1,000 & 45,000  & ft     \\
Target velocity (\texttt{vel\_tgt})         &  300  & 700              & kt    \\
Target heading (\texttt{hdg\_tgt})          &  -180 & 180               & deg  \\
Target off-boresight (\texttt{rgt\_tgt})    &  -60  & 60                 & deg  \\
Distance shooter-target (\texttt{dist})     &  5  & 45          & NM    \\
Target response time (\texttt{delay})       &  15 & 30             & s  \\
Target turn degree (\texttt{turn\_dg})      &  0 & 180              & deg  \\
Target acceleration (\texttt{load\_factor}) &  1.5 & 5             & G (m/s\textsuperscript{2})    \\
\hline
\end{tabular}
\end{table}

Since the LHS method allows the user to specify the desired sample points, we created 10 million different input files. The authors used this work as an opportunity to thoroughly test the simulator and work with a high-fidelity model, considering the large number of variations in input parameters and the potential for non-realistic combinations to arise. Additionally, previous works by the authors using the same simulator provided indications of the necessity of a large volume of data to achieve reliable and meaningful results. These 10 million input files generated an equal number of missile launch simulations, which were executed using an Intel Xeon Silver 4210R CPU with 2.40GHz and 128 GB of RAM. It took approximately 40 days to execute all the simulation runs. Each run generated an output file containing the missile's miss distance (\texttt{miss\_dist}), measured in meters by the simulator, under the respective input conditions. The miss distance is measured at the closest point of approach to the target aircraft and represents the minimum separation distance between the missile and its target~\citep{ball2003fundamentals}. The large number of simulations also allows for an effective downsampling process, as described in the subsection \ref{subsec:preprocessing}, ensuring that a representative subset can be selected while maintaining the fidelity of the original data.

While the LHS method ensures a well-distributed sample across the input parameter space, the simulation data generated in this study is subject to certain limitations due to assumptions and simplifications inherent in the model. One such limitation is the idealized behavior of both the shooter and target, which may not fully capture the complexity of real-world engagements, such as pilot decision-making under uncertainty or environmental factors like varying weather conditions. Additionally, some parameters, such as missile dynamics, were modeled based on a consistent structure, but without accounting for potential variations in performance due to system degradation or other operational constraints.

The simplifications in target response and maneuvering behavior might also limit the accuracy of the model when applied to diverse combat scenarios. Moreover, the computational resources limited us to certain ranges and resolutions of parameters, which might not entirely cover all possible operational scenarios. These factors must be taken into account when interpreting the results, as they could impact the generalizability of the findings to different environments and conditions. Despite these limitations, the insights gained from this analysis provide a valuable foundation, which could be further tested and validated using more complex scenarios and comprehensive simulation environments.

\subsection{Preprocessing}
\label{subsec:preprocessing}

Concerning the preprocessing step, we initially performed an Exploratory Data Analysis (EDA) to identify the dataset's general behaviors and obtain an introductory understanding of the data. This understanding allowed us to formulate hypotheses and assess the need for new data collections. The methods employed for this primary analysis included descriptive statistics, histograms, box plots, correlation, and outlier analysis.

In our study, we employed histograms and box plots to understand the output variable's distribution and behavior across simulations and identify initial outliers in the samples. A crucial aspect of our analysis was the investigation of correlations among variables to detect multicollinearity, which can significantly impair the efficacy of supervised machine learning algorithms~\citep{alin2010multicollinearity}. Notably, we observed a substantial correlation between the Target turn degree (\texttt{turn\_dg}) and Target acceleration (\texttt{load\_factor}). This finding is particularly relevant, suggesting the redundancy of one of these variables. Therefore, we propose eliminating \texttt{load\_factor} from the set of input variables to optimize the performance of our future machine learning models.

Furthermore, we employed a downsampling technique on the initial LHS design to examine potential outliers in the data collected. Identifying outliers and inadequate data in a dataset is arguably one of the most challenging parts of the preprocessing stage, and it is always a topic to be explored cautiously~\citep{kazil2016data}. Based on subject matter expert knowledge, we defined the variable intervals but did not control the combination of values generated for each parameter. As a result, we could produce some improbable input variables. For instance, an aircraft at 1,000 ft launching a missile at a target at 45,000 ft is extremely rare from an operational perspective since a pilot would most likely increase altitude before shooting~\citep{dantas2021weapon}. We manually removed these disfavored samples, such as the one presented, to prevent confusion in the model. For more details, check the code associated with the EDA methods~\footnote{EDA code: \url{https://github.com/jpadantas/poker/tree/main/eda}}.

\subsection{Warhead Lethality}
\label{sub:warhead-lethality}

The lethality of a proximity-fuzed warhead, represented by the probability of kill ($P_{\text{kill}}$), depends on the conditions at the time of detonation and the relative position between the warhead and the target aircraft. The $P_{\text{kill}}$ function estimates the likelihood that an aircraft is destroyed given the detonation of the warhead~\citep{ball2003fundamentals}. This probability is influenced by several factors, including fragment velocity, density of fragment dispersion, and the miss distance.

\begin{figure}[ht]
\centering
\includegraphics[width=\linewidth, trim={0.5cm 0cm 0.5cm 0cm}, clip]{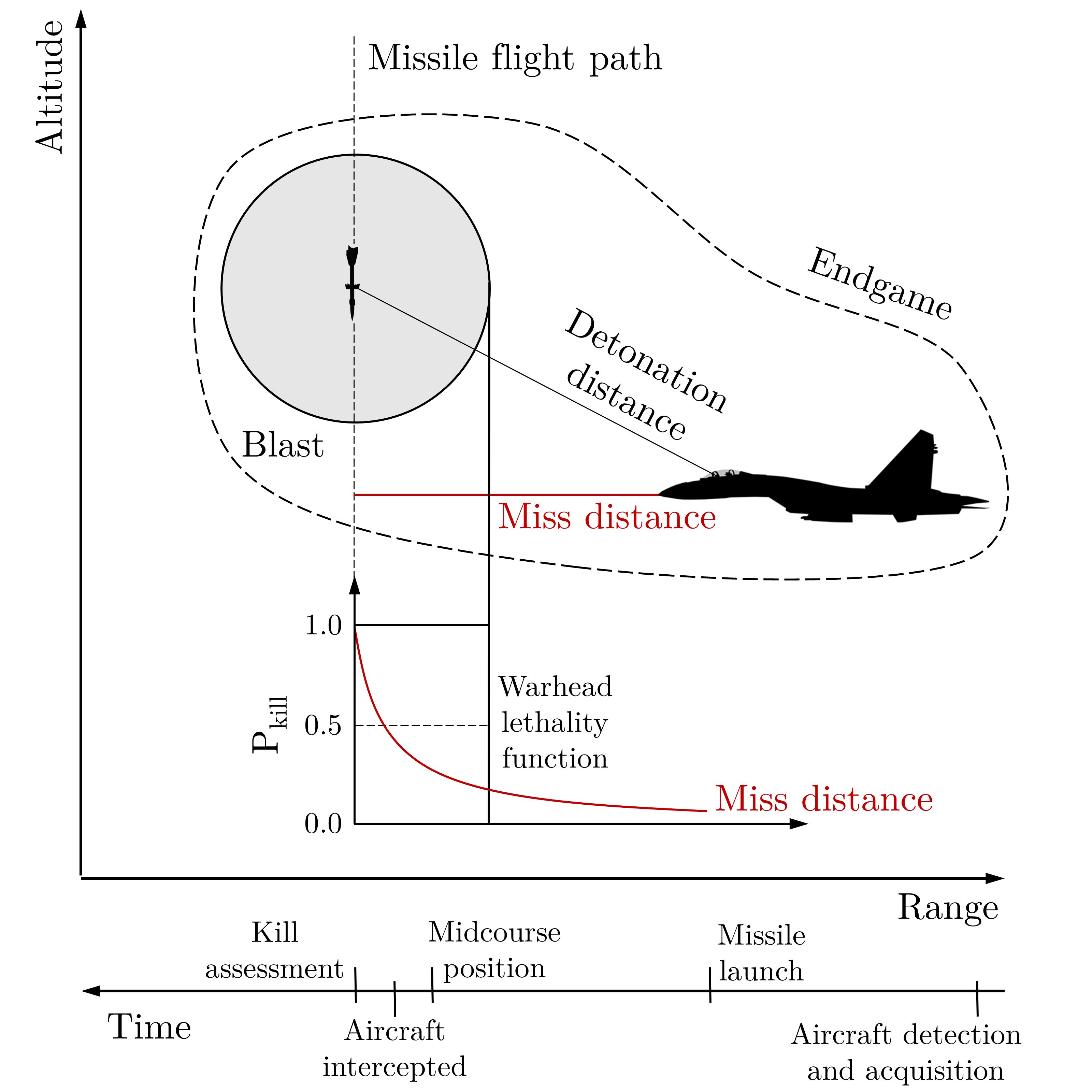}
\caption{Impact of Detonation Distance on Warhead Lethality. The figure shows the relationship between detonation distance and $P_{\text{kill}}$ for a proximity-fuzed warhead. The lethal radius represents the optimal distance for maximum lethality, contrasted with the decreasing kill probability as the distance increases.}
\label{fig: missile}
\end{figure}

Figure~\ref{fig: missile} illustrates how increasing the detonation distance reduces the effectiveness of the warhead. The ideal lethal radius represents the optimal distance for maximum lethality, while the declining effectiveness curve shows how the kill probability decreases as the detonation distance deviates from this optimal value~\citep{ball2003fundamentals}.

\subsubsection{Factors Affecting Warhead Lethality}

The lethality of a proximity-fuzed warhead is determined by various factors that influence how effectively the warhead fragments impact the target. Key factors include:

\begin{itemize}

  \item \textbf{Miss Distance}: The proximity of the warhead at detonation to the target is the primary factor affecting the probability of kill. A smaller miss distance generally leads to a higher likelihood of target destruction, while a larger miss distance reduces effectiveness.

  \item \textbf{Fragment Distribution and Density}: The arrangement and density of the fragments around the explosive affect how many fragments are likely to impact the target. In this study, the pre-formed fragments are arranged in a brickwork pattern, which optimizes coverage and ensures a consistent fragment distribution.

  \item \textbf{Fragment Initial Velocity}: The initial velocity of the fragments post-detonation, which depends on the Gurney constant and the explosive characteristics, determines how far the fragments travel and their ability to penetrate the target structure.

  \item \textbf{Warhead and Explosive Properties}: The material composition, dimensions, and explosive characteristics (e.g., detonation velocity, explosive mass) collectively impact the lethality.

\end{itemize}

\subsubsection{Modeling the Physical Properties of the Warhead}

To calculate $P_{\text{kill}}$, it is necessary to accurately model the warhead's physical properties. Table~\ref{tab:warhead-data} presents the structural, fragment, and explosive properties of the warhead used in this study.

\begin{table}[ht]
\scriptsize
\centering
\caption{Missile's warhead data including structural, fragment, and explosive properties.}
\label{tab:warhead-data}
\begin{tabular}{|c|c|c|}
\hline
\textbf{Variable}                 & \textbf{Value}  & \textbf{Unit} \\ \hline
\multicolumn{3}{|c|}{\textbf{Explosive Data}} \\ \hline
Explosive Type                        & PBX-9404        & -             \\ \hline
Explosive Density ($\rho_{\text{e}}$) & 1710           & $\text{kg/m}^3$      \\ \hline
Explosive Mass ($M_{\text{e}}$)       & 8.94            & kg            \\ \hline
Detonation Velocity                   & 9500           & m/s          \\ \hline
Gurney Constant ($G$)                 & 2895           & m/s          \\ \hline
Initiation Point ($X_2$)              & 100             & mm            \\ \hline
Charge-to-Mass Ratio ($C/M$)          & 0.65            & -             \\ \hline
Fragment Initial Velocity ($V_0$)     & 2028           & m/s             \\ \hline
\multicolumn{3}{|c|}{\textbf{Material Data}} \\ \hline
Material                              & Mild Steel      & -             \\ \hline
Density ($\rho$)                      & 7850           & $\text{kg/m}^3$      \\ \hline
\multicolumn{3}{|c|}{\textbf{Geometric Data}} \\ \hline
External Diameter                     & 178             & mm            \\ \hline
Skin Thickness                        & 2               & mm            \\ \hline
Structural Diameter ($D_{\text{s}}$)  & 154             & mm            \\ \hline
Internal Diameter ($D_{\text{i}}$)    & 144             & mm            \\ \hline
Structural Thickness ($T_{\text{s}}$) & 5               & mm            \\ \hline
Effective Length ($L_{\text{e}}$)     & 321             & mm            \\ \hline
Structural Mass ($M_{\text{s}}$)      & 5.90            & kg            \\ \hline
\multicolumn{3}{|c|}{\textbf{Pre-formed Fragment Data}} \\ \hline
Type                                  & Sphere          & -             \\ \hline
Pattern                               & Brickwork       & -             \\ \hline
Diameter or Distance ($\overline{\text{AO}}$)  & 10              & mm            \\ \hline
Distance ($\overline{\text{OB}}$)              & 8.66            & mm            \\ \hline
Sphere Mass ($m_{\text{sph}}$)               & 0.0041          & kg            \\ \hline
\multicolumn{3}{|c|}{\textbf{Number of Spheres}} \\ \hline
Total Number ($N_{\text{sph}}$)       & 1909            & -             \\ \hline
Total Mass of Spheres ($M_{\text{sph}}$)  & 7.85            & kg            \\ \hline
\end{tabular}
\end{table}

The properties presented in Table~\ref{tab:warhead-data} form the foundation for understanding the warhead's behavior during detonation, which is essential for modeling its effectiveness. Each key parameter contributes to the warhead's ability to distribute fragments and ultimately destroy the target. The data is organized as follows:

\begin{itemize}

  \item \textbf{Explosive Data}: The warhead uses PBX-9404, a high-energy explosive with a density ($\rho_{\text{e}}$) and detonation velocity ($D$). The explosive mass ($M_{\text{e}}$), Gurney constant ($G$), and the initiation point ($X_2$) are key factors in determining the velocity and energy imparted to the fragments upon detonation.

  \item \textbf{Material Data}: The warhead casing is made from mild steel, providing sufficient strength to contain the explosive until detonation. The density ($\rho$) of the steel is used to calculate structural mass and other characteristics relevant to energy transfer during detonation.

  \item \textbf{Geometric Data}: Geometric properties such as structural diameter ($D_{\text{s}}$), internal diameter ($D_{\text{i}}$), and effective length ($L_{\text{e}}$) determine the size and volume of both the warhead structure and the explosive content. These dimensions influence explosive power and fragment distribution.

  \item \textbf{Pre-formed Fragment Data}: The fragments are pre-formed as spherical mild steel pieces and arranged in a brickwork pattern. This arrangement ensures uniform fragment dispersion. Parameters include fragment diameter ($\overline{\text{AO}}$), distance between fragment centers ($\overline{\text{OB}}$), and sphere mass ($m_{\text{sph}}$). The total number of fragments ($N_{\text{sph}}$) and their combined mass ($M_{\text{sph}}$) are essential for assessing lethality.

  \item \textbf{Fragment Initial Velocity ($V_0$)}: Using the Gurney equation, the initial velocity of the fragments ($V_0$) is estimated. This velocity is critical for determining how far and how fast the fragments will travel toward the target, which directly impacts $P_{\text{kill}}$.

\end{itemize}

In summary, the combination of explosive, material, geometric, and fragment properties forms the foundation of the warhead's lethality model. These data points will be integrated into subsequent probabilistic models to evaluate $P_{\text{kill}}$ under varying combat scenarios, providing a comprehensive assessment of the warhead's effectiveness.

\subsubsection{Warhead Performance Metrics}

The initial velocity ($V_0$) of the warhead fragments post-detonation is determined using the Gurney equation:

\begin{equation}
\label{eq:8}
V_0 = G\cdot\sqrt{\frac{C/M}{1+0.5 \cdot C/M}},
\end{equation}

\noindent{where $G$ is the Gurney constant of the explosive, and $C/M$ is the ratio of the explosive mass to the total mass of the warhead's structural components and fragments.}

Additionally, static ejection angles are necessary for determining the initial dispersion of the fragments. These angles represent the directional pattern of fragment ejection relative to the warhead, influenced by the warhead's design and explosive-structure interaction. Understanding these angles helps optimize fragment spread for maximum target coverage.

\subsubsection{Endgame Condition Determination}

In this stage, the model considers operational flight conditions at the moment of detonation. The conditions include:

\begin{itemize}

\item \textbf{Missile Velocity ($V_m$)}: Determines the additional kinetic energy imparted to the fragments, influencing their relative velocity to the target.
\item \textbf{Target Velocity ($V_t$)}: Affects the relative speed and encounter geometry, impacting the likelihood of fragment impact.
\item \textbf{Miss Distance}: The predicted closest distance between the missile and the target at detonation, directly affecting $P_{\text{kill}}$ by determining how many fragments can reach the target.
\item \textbf{Altitude}: Influences air density, affecting drag on the fragments and their ability to penetrate the target.

\end{itemize}

\subsubsection{Endgame Dynamics Computation}

The endgame dynamics involve computing dynamic ejection angles and fragment velocities relative to the missile's trajectory to predict the interaction with the target. Unlike static ejection angles, dynamic ejection angles consider missile velocity at detonation, modifying the trajectory of each fragment.

The relative velocity of each fragment with respect to the moving target is calculated to assess the likelihood of impact. This velocity depends on the initial velocity of the fragment ($V_0$), missile velocity ($V_m$), and target velocity ($V_t$). The computed relative velocity and ejection angles allow for the estimation of the distance each fragment will travel post-detonation, which is crucial for assessing their ability to reach and penetrate the target.

\subsubsection{Fragment Spray Density and Lethality Assessment}

The fragment spray density represents the concentration of fragments within the target zone, a key factor in determining lethality. The density of fragment saturation directly influences the number of impacts on the target, affecting $P_{\text{kill}}$.

Several factors are combined to evaluate lethality, including:

\begin{itemize}

\item \textbf{Drag Coefficient ($C_D$)} and \textbf{Air Density ($\rho$)}: These parameters affect how quickly fragment velocity decreases due to air resistance. A higher air density or drag coefficient results in greater velocity reduction, reducing penetration likelihood.

\item \textbf{Impact Velocity}: The impact velocity of the fragments determines their ability to inflict damage upon reaching the target. This velocity is influenced by the drag experienced during flight.

\end{itemize}

The lethality assessment combines these factors to estimate the probability of kill per fragment hit, ultimately leading to the overall kill probability for each fragment.

\subsubsection{Computation of Final Probability of Kill}

For the calculation of $P_{\text{kill}}$ in this work, it was assumed that all missile launches used consistent values for key parameters. The intrinsic missile parameters, which are assumed to remain constant across different simulations, include: fragment initial velocity $V_0 = 2028$ m/s, effective length $L_{\text{e}} = 321$ mm, internal diameter $D_{\text{i}} = 144$ mm, structural diameter $D_{\text{s}} = 154$ mm, and fragment diameter $\overline{\text{AO}} = 10$ mm.

On the other hand, the parameters that may vary depending on the simulation context are the missile velocity $V_m = 592$ m/s, target velocity $V_t = 250$ m/s, target wingspan of 8.13 m, and the angle between trajectories, which was set to 0 radians for this study. This distinction helps clarify which parameters are intrinsic to the missile type and which are subject to change depending on the specific conditions of the simulation.

Since these parameters remain constant for all launches, the miss distance becomes the primary factor influencing the variation in $P_{\text{kill}}$. This assumption simplifies the model, allowing us to focus on optimizing warhead lethality based on detonation distance.

The final $P_{\text{kill}}$ integrates the structural and explosive characteristics of the warhead with dynamic ejection analysis and fragment lethality assessment. This probability provides a statistical measure of the warhead's effectiveness and its likelihood of achieving its intended destructive outcome upon the target.

This model integrates principles from materials science, ballistics, aerodynamics, and probability theory to predict warhead performance, from detonation to target impact. By incorporating the miss distance obtained from the simulation, we transform this distance into a $P_{\text{kill}}$ value, enhancing the model's predictive capacity. Such an approach is important for optimizing warhead configurations to maximize their efficacy. Interested readers can refer to the code for lethality calculations available at the following repository~\footnote{Lethality code: \url{https://github.com/jpadantas/poker/tree/main/lethality}}.

\subsection{Supervised Machine Learning Models}

We created supervised machine learning models using a subset of the initially proposed input variables, excluding the \texttt{load\_factor} variable, as mentioned previously, to develop a regression model capable of predicting the missile $P_{\text{kill}}$, which was calculated from the dataset based on the miss distance from the target, using the lethality theory described in the previous subsection. Notably, variables such as \texttt{turn\_dg} and \texttt{delay} are not directly available to the launcher and require assumptions for their estimation. These variables will be addressed in detail in a later subsection. We employed three distinct algorithms to build the models: Polynomial Regression (PR), Artificial Neural Networks (ANN), and Extreme Gradient Boosting (XGBoost).

Before starting creating and training the models, we performed a stratified train-validate-test split for all supervised machine learning algorithms regarding 34 different probability blocks (bins) using Doane's formula~\citep{doane1976aesthetic}, allocating 80\% for training and validation using a 5-fold cross-validation technique and 20\% for testing, following the approach proposed in previous works~\citep{dantas2021weapon, dantas2023real}. This dataset separation will allow the evaluation of the machine learning models later.

PR extends the linear model by adding extra predictors obtained by raising each original predictor to a determined degree. Adding polynomial terms to the linear model can effectively allow the model to identify nonlinear patterns~\citep{kuhn2019feature}. The degree of the polynomial controls the number of features added. Using a degree greater than four is unusual because the polynomial curve can become overly flexible and take on some unusual shapes~\citep{james2013introduction}. We proposed four different models with degrees from one to four using the Scikit-learn library~\citep{scikit-learn}. For additional information, refer to the code associated with the PR models~\footnote{PR code: \url{https://github.com/jpadantas/poker/tree/main/pr}}.

ANN seeks to approximate the function represented by the dataset, calculating the error between the predicted outputs and the expected outputs and minimizing this error during the training process, working as function approximation machines that are designed to achieve statistical generalization~\citep{goodfellow2017deep}. In this work, we used a multi-layer perceptron using the backpropagation algorithm, adjusting the weights of the connections between ANN neurons to minimize the mean-squared error loss~\citep{reed1999neural}. We performed data scaling to equally distribute the importance of each input in the ANN learning process~\citep{priddy2005artificial}. We used a min-max scaler, transforming all data features to a range from 0 to 1~\citep{bonaccorso2017machine}. For the ANN model, we utilized TensorFlow~\citep{abadi2015tensorflow} to run 50 different hyperparameters configurations, changing the number of hidden layers in \{1, 2, 3, 4, 5, 6, 7, 8, 9, 10\} and the number of units in \{16, 32, 64, 128, 256\}. All nodes have a rectified linear activation function (ReLU)~\citep{goodfellow2017deep}. In addition, we included the Adaptive Moment Estimation (Adam) optimizer, a well-known training algorithm for ANN~\citep{bock2019proof}. Adam is a stochastic gradient descent method based on adaptive estimation of first- and second-order moments function~\citep{kingma2014adam}. We typically should choose a batch size between one and a few hundred. For a given computational cost, small batch sizes achieve the best training stability and generalization performance across diverse experiments~\citep{masters2018revisiting}. We selected 32 as the model's batch size, considered a reasonable default value~\citep{bengio2012practical}. For more details, check the code associated with the ANN models~\footnote{ANN code: \url{https://github.com/jpadantas/poker/tree/main/ann}}.

XGBoost, a leading gradient boosting framework, is engineered to construct a robust ensemble of decision trees by iteratively correcting errors from previous trees, thereby enhancing the model's accuracy for regression, classification, and ranking tasks and is considered the state-of-the-art for tabular data, as evidenced by its widespread adoption and superior performance in various competitions and applications~\citep{chen2016xgboost}. In our study, we leveraged an extensive array of hyperparameters, including \texttt{n\_estimators} set to an ambitious 1,000,000 trees for comprehensive data exploration, and \texttt{max\_depth} options ranging in \{10, 12, 14, 16, 18, 20\}, enabling the model to uncover complex patterns without excessively increasing complexity. The learning rate was varied across \{0.01, 0.1, 0.2, 0.3, 0.4\}, balancing the convergence speed with the solution's accuracy. Alongside this, subsampling and feature sampling rates were set using \texttt{subsample}, \texttt{colsample\_bytree}, and \texttt{colsample\_bylevel}, which were varied in the range from 0.5 to 1.0 with a step size of 0.1, to enhance model generalization. Additionally, \texttt{min\_child\_weight} and \texttt{gamma} parameters were finely adjusted to regulate the model's growth and complexity, with \texttt{min\_child\_weight} in \{1, 3, 5\} and \texttt{gamma} in \{0.0, 0.1, 0.2, 0.3, 0.4\}. For further insights, please consult the code for the XGboost models~\footnote{XGBoost code: \url{https://github.com/jpadantas/poker/tree/main/xgboost}}.

For both the ANN and XGBoost models, we implemented an early-stopping mechanism to oversee the training phase efficiently. This strategy entailed continuous monitoring of performance metrics on the validation set to detect any stagnation in improvement. We defined ``patience'' as the maximum number of epochs to continue training without observing any advancement in the validation set's metrics, setting this threshold at 10 epochs. This specific interval proved to be balanced for mitigating the effects of noise within the model optimization processes, ensuring that both models ceased training at a juncture that prevented overfitting while maximizing performance~\citep{goodfellow2017deep}.

The choice of machine learning models in this study was made to provide a balanced comparison between simplicity and performance. We included PR as a simpler baseline model, which we anticipated would be faster to train but likely result in lower performance compared to more complex models. This choice allowed us to understand the trade-offs between model complexity and accuracy in our context. The ANN and XGBoost models were selected as they are well-known for their effectiveness with tabular data and are frequently used for achieving high predictive metrics in such scenarios. This combination of models ensured a comprehensive evaluation of different approaches, ranging from simpler to more sophisticated techniques, thereby providing valuable insights into model performance under different levels of complexity.

\subsection{Models Evaluation}

Since this work analyzed a regression problem, we evaluated all the supervised machine learning models using the following well-known metrics: Mean Absolute Error (MAE), Mean Squared Error (MSE), Root Mean Squared Error (RMSE), and the Coefficient of Determination (R²). Additionally, training and inference time are two critical aspects of a model's assessment. We analyzed each cross-validation fold's training time and calculated each model's inference time on the test dataset. While training time is important, inference time is even more important as it directly impacts the model's practical applicability in real-time scenarios. We can estimate which model offers the best trade-off between performance and computational cost by examining these factors.

\subsection{Estimating Turn Degree and Delay}
\label{subsec:turn-degree-and-delay}

Estimating the \texttt{turn\_dg} and \texttt{delay} for a model presents significant challenges due to the inherent unpredictability of a target's response to an incoming missile. This unpredictability is largely attributed to the unique characteristics and decision-making processes of the target, which can vary widely in real-world scenarios. Understanding and predicting how a target will maneuver in response to a perceived threat is critical for accurate simulation and assessment of missile engagement outcomes.

To address this challenge, we employ the expertise of SMEs in military tactics and missile defense systems to estimate the potential \texttt{turn\_dg} and \texttt{delay} of a target upon missile detection. The approach leverages the deep understanding of SMEs regarding the behavior of various targets under threat, allowing for a more informed and realistic set of estimations.

These parameters are essential for modeling the target's evasive maneuvers and the subsequent trajectory adjustments of the missile. We represent the target's response variability using normal distributions for both \texttt{turn\_dg} and \texttt{delay}. The choice of a normal distribution is based on the assumption that the target's response will vary around a mean value (median) with a certain degree of predictability, as influenced by the target's characteristics and situation. These distributions' median and standard deviation are determined through a collaborative process with SMEs, ensuring that the estimations reflect a realistic range of target behaviors.

The estimation process involves the identification of specific scenarios in which a target might find itself when a missile is launched toward it. Each scenario is carefully analyzed to determine the likely response of the target, focusing on two key parameters: the \texttt{turn\_dg} and \texttt{delay} before initiating the turn.

To estimate the \texttt{turn\_dg} of the target, we assume that the target will adopt the most defensive posture possible, always attempting to move in the direction opposite to the missile launcher. Regardless of the distance between the missile launcher and the target, the target aircraft will aim to maximize the separation by performing a defensive turn of up to 180 degrees, whenever feasible. This approach ensures that the target consistently prioritizes evasion by moving away from the threat to the greatest extent possible.

The initial value of the target's turn degree is defined by the following function:

\begin{equation}
\label{eq:2}
\scalebox{1}{$\texttt{turn\_dg}_0 = \left( (\texttt{rgt\_tgt} + 180) \bmod 360 \right) - 180$}.
\end{equation}

\begin{figure}[ht]
    \centering
    \includegraphics[width=\linewidth, trim={2cm 2cm 2cm 2cm}, clip]{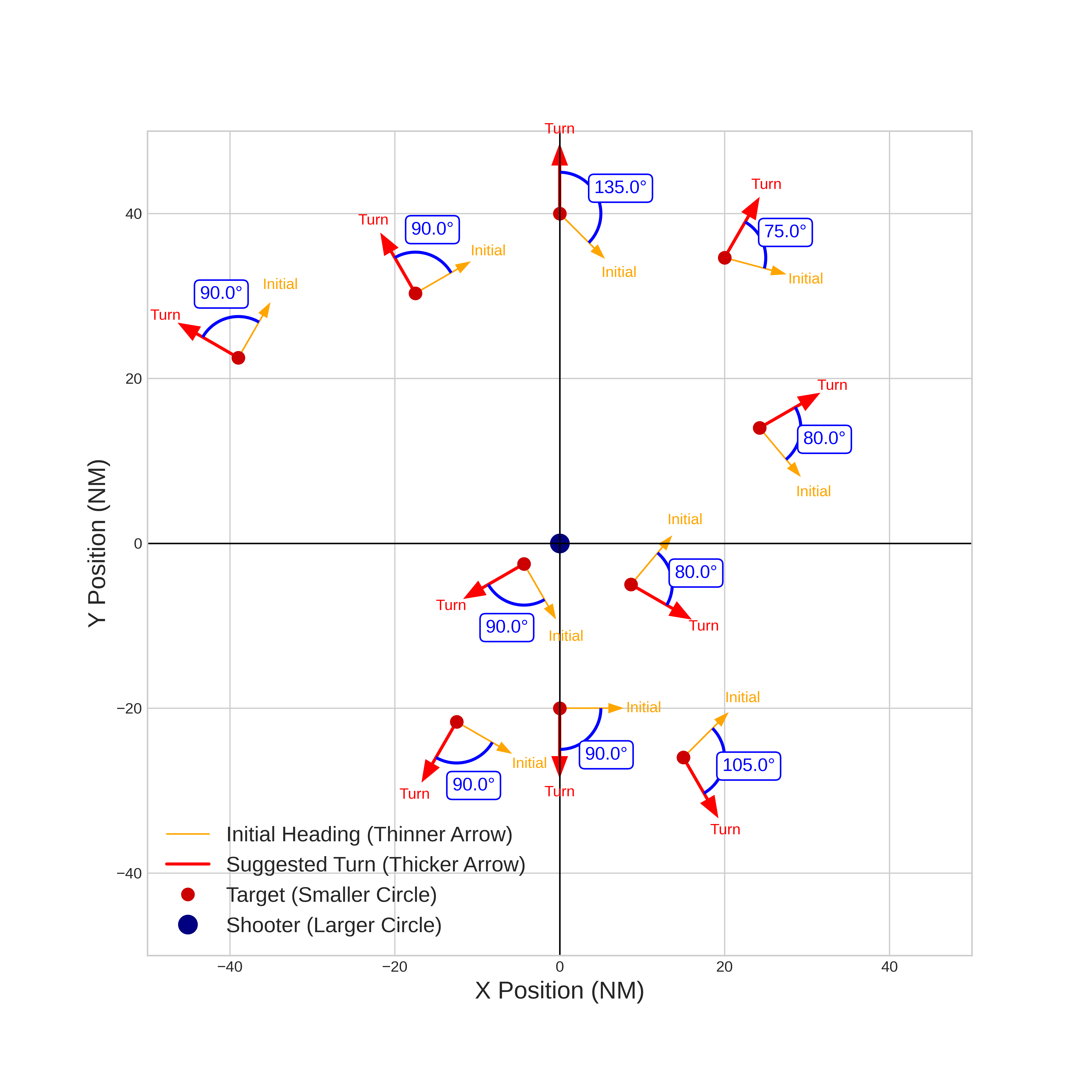}
    \caption{Illustration of ten diverse scenarios of defensive turn maneuvers. Each scenario shows a target attempting to maximize distance from the launcher. The dark red points (smaller) indicate the initial position of the target, while the orange (thinner) and red arrows (thicker) represent the initial heading and the suggested turn direction, respectively. The dark circle (larger) represents the shooter. The blue arc indicates the turn degree (\texttt{turn\_dg}) performed by the target, which varies between 0° and 180°.}
    \label{fig:defensive_turn_maneuvers}
\end{figure}

To illustrate the initial value of the defensive behavior of the targets, Figure~\ref{fig:defensive_turn_maneuvers} presents ten diverse and well-spread scenarios, where each target attempts to perform a defensive maneuver aimed at maximizing separation from the launcher. The distances, initial headings, and off-boresight angles were chosen to ensure a good distribution in space, avoiding overlap and facilitating visualization of the maneuvers. Each target, represented by the red point, performs a turn to move away from the launcher (the blue point at the center) according to the suggested direction indicated by the red arrow, in an effort to maximize evasion. The turn degree ($\texttt{turn\_dg}_0$) is represented by the blue arc and varies according to the relative position of the target to the launcher.

To better reflect realistic behavior, the final value of \texttt{turn\_dg} is modeled with added variability as follows:

\begin{equation}
\scalebox{1}{$
\texttt{turn\_dg} = \texttt{turn\_dg}_0 \left( 1 + 0.05 \cdot w \right)
$},
\end{equation}

\noindent{where \(\texttt{turn\_dg}_0\) is the value calculated from Equation~\ref{eq:2}, and \(w \sim \mathcal{N}(0, 1)\) represents a standard normal random variable. This introduces a 5\% variation, which accounts for natural fluctuations in the target's maneuver.}

Moreover, regarding the variable \texttt{delay}, we anticipate specific behaviors based on the target's orientation relative to the missile launcher. While we initially assumed that shorter reaction times would occur when the target directly faces the launcher due to the immediate perception of the threat, we acknowledge that this is not always the case. The reaction time depends on various factors, including the target's sensor capabilities and engagement strategy. For instance, even when the launcher is positioned behind the target, a Radar Warning Receiver (RWR) may immediately detect the radar, leading to a prompt response.

\begin{figure}[ht]
    \centering
    \includegraphics[width=\linewidth, trim={0.3cm 0.3cm 0.3cm 0.3cm}, clip]{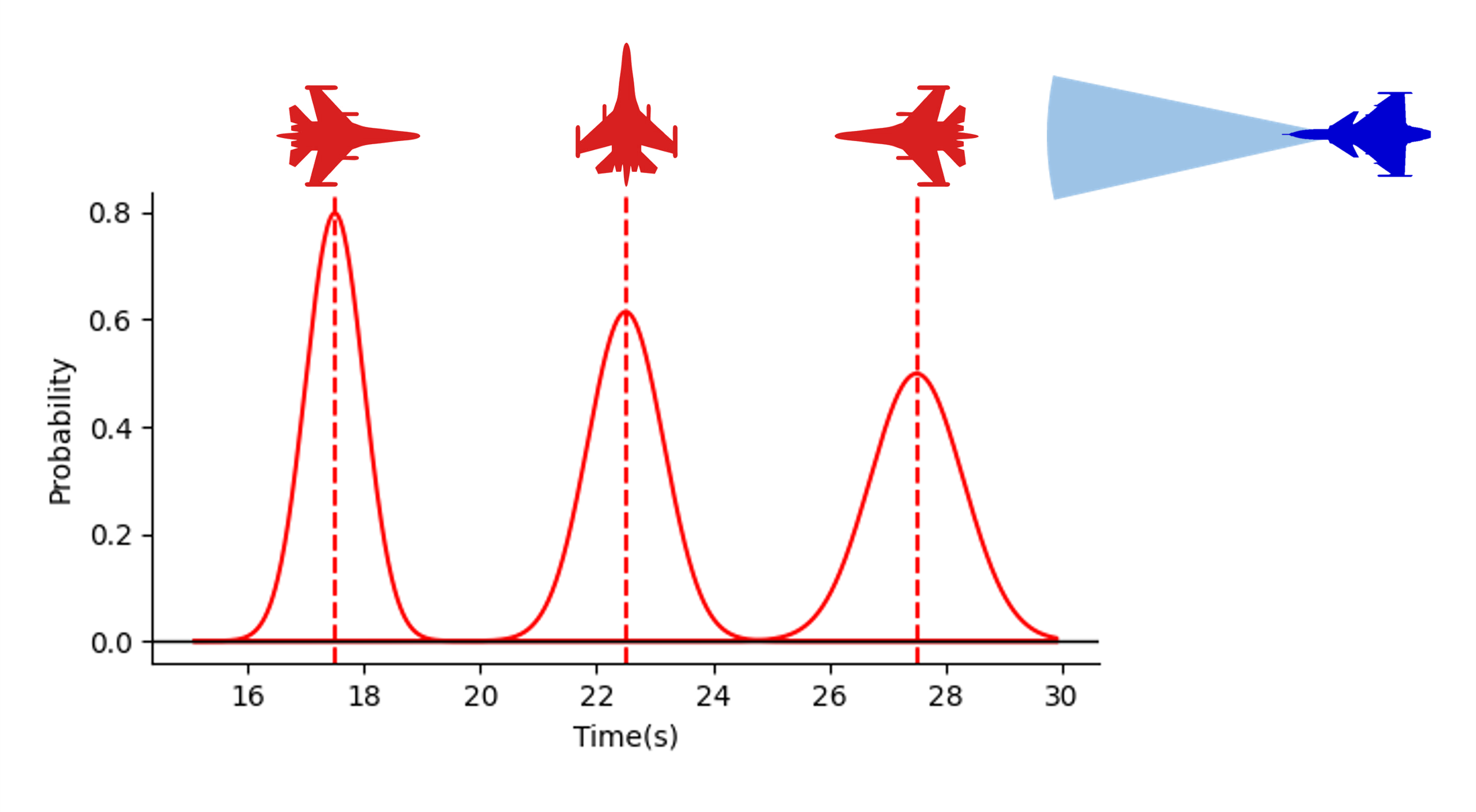}
\caption{Graph illustrating the target's delay time as a function of its orientation with respect to the shooter.}
    \label{fig:delay}
\end{figure}

To simplify this context, we made certain assumptions about the target's behavior. Specifically, we structured the estimated reaction times within a range of 15 to 30 seconds. This range is further segmented into three distinct intervals, each spanning 5 seconds, to reflect varying degrees of threat awareness and response times. These intervals are divided into 120 degrees each to cover the full 360-degree range, ensuring that all possible orientations are accounted for. Each interval is associated with a standard deviation value of approximately 5\%, which ensures the three curves are almost connected, allowing for comprehensive sampling of all potential values. Please refer to Figure~\ref{fig:delay} for a detailed visualization of the considered scenarios and the corresponding distributions for \texttt{delay}.

To validate the chosen values for the median and standard deviation, we employ the face validation method~\citep{balci1998verification}. This method involves a review by an independent group of experts who assess the plausibility and accuracy of the estimations based on their own experience and knowledge. The face validation process ensures that the estimations are not only based on a solid theoretical foundation but also resonate with the practical insights of seasoned professionals in the field.

It is important to note that the parameters and distributions outlined in this section adhere to the intervals and assumptions previously described in this work. By leveraging SME knowledge and employing rigorous validation techniques, we aim to provide a robust framework for estimating \texttt{turn\_dg} and \texttt{delay}, thereby enhancing the accuracy of missile engagement simulations. The validation process was conducted through face validation by SMEs with extensive experience in missile engagements. Specifically, several simulated scenarios were presented to the SMEs, covering a range of engagement distances, target maneuvers, and missile launches. The SMEs evaluated whether the estimated \texttt{turn\_dg} and \texttt{delay} values aligned with their operational experience and expectations under different combat situations. This process ensured that the estimated values were realistic and consistent with real-world operational behavior.

\subsection{The Probability of Hitting a Target}

Determining the final probability of hitting a target is intricately linked to the analysis performed by machine learning algorithms. These algorithms are designed to process a comprehensive set of input variables derived from the aircraft's subsystems, providing the pilot with actionable intelligence to enhance missile targeting accuracy. However, a critical component of this targeting process is the estimation of the target's \texttt{turn\_dg} and \texttt{delay}, which are not directly observable and must be inferred through the methodology outlined previously.

This estimation plays an essential role in calculating the hit probability, accounting for the variability in the target's response behavior. By incorporating the target's expected maneuvers, the machine learning model can adjust its predictions to better reflect the dynamic nature of aerial engagements. The aim is to achieve a conservative estimation of the target's performance, ensuring that the probability of a hit considers the best possible maneuvers the target could execute to evade the missile.

To this end, the model leverages the variability in the target's behavior as an essential factor, allowing for a better understanding of potential evasive actions. This approach is grounded in the principle of anticipating optimal target performance, where the target utilizes its capabilities to the maximum extent to evade incoming threats. By assuming a high level of performance from the target, the model inherently considers the target to perform well on average, thereby minimizing the risk of underestimating the target's evasive capabilities.

Integrating this realistic estimation into the machine learning algorithm enhances the reliability of the hit probability calculation. It enables the algorithm to account for the widest possible range of target behaviors, including those that may occur under extreme conditions. This approach ensures the pilot has a probability of hit that reflects the most challenging engagement scenarios, facilitating informed decision-making during missile launch sequences.

In summary, the probability of hitting a target is not merely a static figure derived from straightforward inputs; it is a dynamic estimation that incorporates the complexity and unpredictability of aerial combat. By integrating conservative estimations of the target's \texttt{turn\_dg} and \texttt{delay} into the machine learning response, we aim to improve the accuracy and realism of missile engagement outcomes. However, we acknowledge the limitations of the current approach, and further empirical validation is required to fully demonstrate the practical benefits of these estimations. This ongoing work lays the foundation for future studies that will more comprehensively assess the impact of these estimations on aerial defense strategies.
\section{Results and Analysis}
\label{results}

This section examines the metrics of the test dataset for the proposed machine learning models. Additionally, we provide a Multi-Function Display (MFD) representation, incorporating the traditional maximum range and no-escape zone indicators along with the probability of kill in the WEZ indication based on the proposed probabilistic model.

\subsection{Comparative Analysis of the Machine Learning Models}

Table~\ref{tab:metrics} summarizes the mean performance metrics for each model across the five folds of the three distinct proposed models: PR, ANN, and XGBoost. These metrics include MAE, MSE, RMSE, and R², reflecting their respective best hyperparameter settings, chosen based on performance on the validation dataset during training. It is important to note that the predicted variable, \( P_{\text{kill}} \), is expressed as a percentage. Consequently, all the error metrics (MAE, MSE, RMSE) are dimensionless, as they are derived from percentage values. Additionally, R² is also dimensionless. The analysis further includes computational efficiency metrics, such as training and inference time, both measured in seconds, using the training/validation and test datasets, respectively.

\begin{table}[htb]
\scriptsize
\centering
\caption{Comparative evaluation of machine learning models based on the proposed metrics. The best results for each metric are highlighted in bold red text.}
\label{tab:metrics}
\begin{tabular}{|c|c|c|c|}
\hline
\textbf{Model} & \textbf{PR} & \textbf{ANN} & \textbf{XGBoost} \\ \hline
\makecell{\textbf{Hyperparameters}} &
\makecell{degree=4} &
\makecell{hidden\_layers=3,\\ units=128} &
\makecell{max\_depth=14,\\ learning\_rate=0.01,\\ subsample=0.9,\\ colsample\_bytree=0.9,\\ colsample\_bylevel=0.6,\\ min\_child\_weight=5,\\ gamma=0.1} \\ \hline
\textbf{MAE} & 11.52 & \textbf{\textcolor{red}{2.33}} & 5.04 \\ \hline
\textbf{MSE} & 296.26 & \textbf{\textcolor{red}{51.90}} & 117.14 \\ \hline
\textbf{RMSE} & 17.21 & \textbf{\textcolor{red}{7.20}} & 10.82 \\ \hline
\textbf{R²} & 67.94 & \textbf{\textcolor{red}{94.38}} & 87.31 \\ \hline
\makecell{\textbf{Training Time (s)}} & \textbf{\textcolor{red}{532.23}} & 4722.17 & 6098.33 \\ \hline
\makecell{\textbf{Inference Time (s)}} & 9.57 & \textbf{\textcolor{red}{6.65}} & 8.20 \\ \hline
\end{tabular}
\end{table}

\subsection{Statistical Analysis}

To determine whether the observed differences in model performance are statistically significant, we performed the Friedman test~\citep{friedman1937use} followed by the Nemenyi post-hoc test~\citep{nemenyi1963distribution} for multiple comparisons. Interested readers can refer to the code for the statistical analysis at the following repository~\footnote{Statistics code: \url{https://github.com/jpadantas/poker/tree/main/statistics}}.

\subsubsection{Friedman Test}

The Friedman test was used to detect differences in performance across the three models using the results from 5 folds of each model. The null hypothesis is that there are no differences in the performance metrics among the models. Table~\ref{tab:friedman_results} presents the results of the Friedman test for each metric.

\begin{table}[htb]
\scriptsize
\centering
\caption{Friedman Test Results. Significant p-values are highlighted in bold red text.}
\label{tab:friedman_results}
\begin{tabular}{|c|c|c|}
\hline
\textbf{Metric} & \textbf{Friedman Statistic} & \textbf{p-value} \\ \hline
MAE             & 10.0     & \textbf{\textcolor{red}{0.0067}}     \\ \hline
MSE             & 10.0     & \textbf{\textcolor{red}{0.0067}}     \\ \hline
RMSE            & 10.0     & \textbf{\textcolor{red}{0.0067}}     \\ \hline
$R^2$           & 10.0     & \textbf{\textcolor{red}{0.0067}}     \\ \hline
Training Time   & 8.4      & \textbf{\textcolor{red}{0.0150}}     \\ \hline
Inference Time  & 4.8      & 0.0907     \\ \hline
\end{tabular}
\end{table}

The results indicate that for all metrics except inference time, there are statistically significant differences among the models (p-value $<$ 0.05). Notably, the p-values for several metrics (MAE, MSE, RMSE, and $R^2$) are identical, suggesting that the models perform similarly across these related metrics. This similarity could be due to the fact that these metrics are strongly correlated, and the relatively small sample size (5 folds) may limit the ability of the Friedman test to distinguish nuanced differences between models. Additionally, the identical Friedman statistics for these metrics imply that the rankings across models were consistent, reflecting similar performance trends.

\subsubsection{Nemenyi Post-hoc Test}

To identify which models differ significantly from each other, we conducted the Nemenyi post-hoc test. Table~\ref{tab:nemenyi} shows the p-values for pairwise comparisons among the models for the metrics where the Friedman test was significant. Note that the inference time metric did not show significant differences in the Friedman test; therefore, the Nemenyi test was not conducted for this metric.

\begin{table}[H]
\scriptsize
\centering
\caption{Nemenyi post-hoc test p-values for pairwise comparisons among models. Significant p-values are highlighted in bold red text.}
\label{tab:nemenyi}
\begin{tabular}{|c|c|c|}
\hline
\textbf{Metric} & \textbf{Comparison} & \textbf{p-value} \\ \hline
\multirow{3}{*}{MAE}
    & PR vs ANN           & \textbf{\textcolor{red}{0.0045}} \\ \cline{2-3}
    & XGBoost vs PR       & 0.2541 \\ \cline{2-3}
    & XGBoost vs ANN      & 0.2541 \\ \hline
\multirow{3}{*}{MSE}
    & PR vs ANN           & \textbf{\textcolor{red}{0.0045}} \\ \cline{2-3}
    & XGBoost vs PR       & 0.2541 \\ \cline{2-3}
    & XGBoost vs ANN      & 0.2541 \\ \hline
\multirow{3}{*}{RMSE}
    & PR vs ANN           & \textbf{\textcolor{red}{0.0045}} \\ \cline{2-3}
    & XGBoost vs PR       & 0.2541 \\ \cline{2-3}
    & XGBoost vs ANN      & 0.2541 \\ \hline
\multirow{3}{*}{$R^2$}
    & PR vs ANN           & \textbf{\textcolor{red}{0.0045}} \\ \cline{2-3}
    & XGBoost vs PR       & 0.2541 \\ \cline{2-3}
    & XGBoost vs ANN      & 0.2541 \\ \hline
\multirow{3}{*}{Training Time}
    & XGBoost vs PR       & \textbf{\textcolor{red}{0.0123}} \\ \cline{2-3}
    & PR vs ANN           & 0.1394 \\ \cline{2-3}
    & XGBoost vs ANN      & 0.6008 \\ \hline
\end{tabular}
\end{table}

The Nemenyi post-hoc test results, shown in Table~\ref{tab:nemenyi}, indicate that significant differences primarily occurred between PR and ANN. For XGBoost, its comparisons with both PR and ANN resulted in similar p-values across multiple metrics. This suggests that XGBoost's performance was relatively stable and consistently fell between the other two models. This pattern might reflect limited differentiation capability given the inherent relationships among the metrics and the modest sample size, highlighting that larger samples might be needed to detect finer performance differences.

\subsection{Discussion}

The statistical analysis confirms that the choice of model has a significant impact on performance metrics and training time. The Friedman test indicated statistically significant differences among the models for MAE, MSE, RMSE, R², and Training Time (p-value $<$ 0.05). The Nemenyi post-hoc test further identified that:

\begin{itemize}
    \item For MAE, MSE, RMSE, and R², ANN significantly outperformed PR (p-value = 0.0045).
    \item No significant differences were found between XGBoost and PR, or XGBoost and ANN for these metrics.
    \item For Training Time, PR had a significantly shorter training time than XGBoost (p-value = 0.0123).
    \item No significant differences were found between PR and ANN, or XGBoost and ANN for Training Time.
\end{itemize}

These results suggest that ANN provides superior predictive accuracy compared to PR, with a statistically significant improvement in error metrics. Although ANN's training time was longer than PR's, the difference was not statistically significant between ANN and PR, indicating that the increased computational cost is acceptable given the performance gains.

Among the models tested, ANN exhibited the lowest error metrics across all considered measures. Specifically, ANN achieved the lowest MAE (2.33), suggesting its predictions were closer to the actual values on average compared to PR and XGBoost. Similarly, ANN recorded the lowest MSE (51.90) and RMSE (7.20), indicating a strong capability to minimize both variance and bias in its predictions. The R² value for ANN stood at 94.38\%, demonstrating that it could explain a substantial portion of the variance in the dataset, which indicates strong predictive power.

Regarding computational efficiency, PR had the shortest training time (532.23 seconds), which was significantly shorter than that of XGBoost (6098.33 seconds) as indicated by the Nemenyi test (p-value = 0.0123). ANN's training time (4722.17 seconds) was not significantly different from either PR or XGBoost. In terms of inference time, ANN was the fastest (6.65 seconds), slightly quicker than XGBoost (8.20 seconds) and PR (9.57 seconds).

\begin{figure*}[ht]
    \centering
    \subfloat[]{
        \includegraphics[width=0.485\textwidth, trim={250pt 180pt 160pt 230pt}, clip]{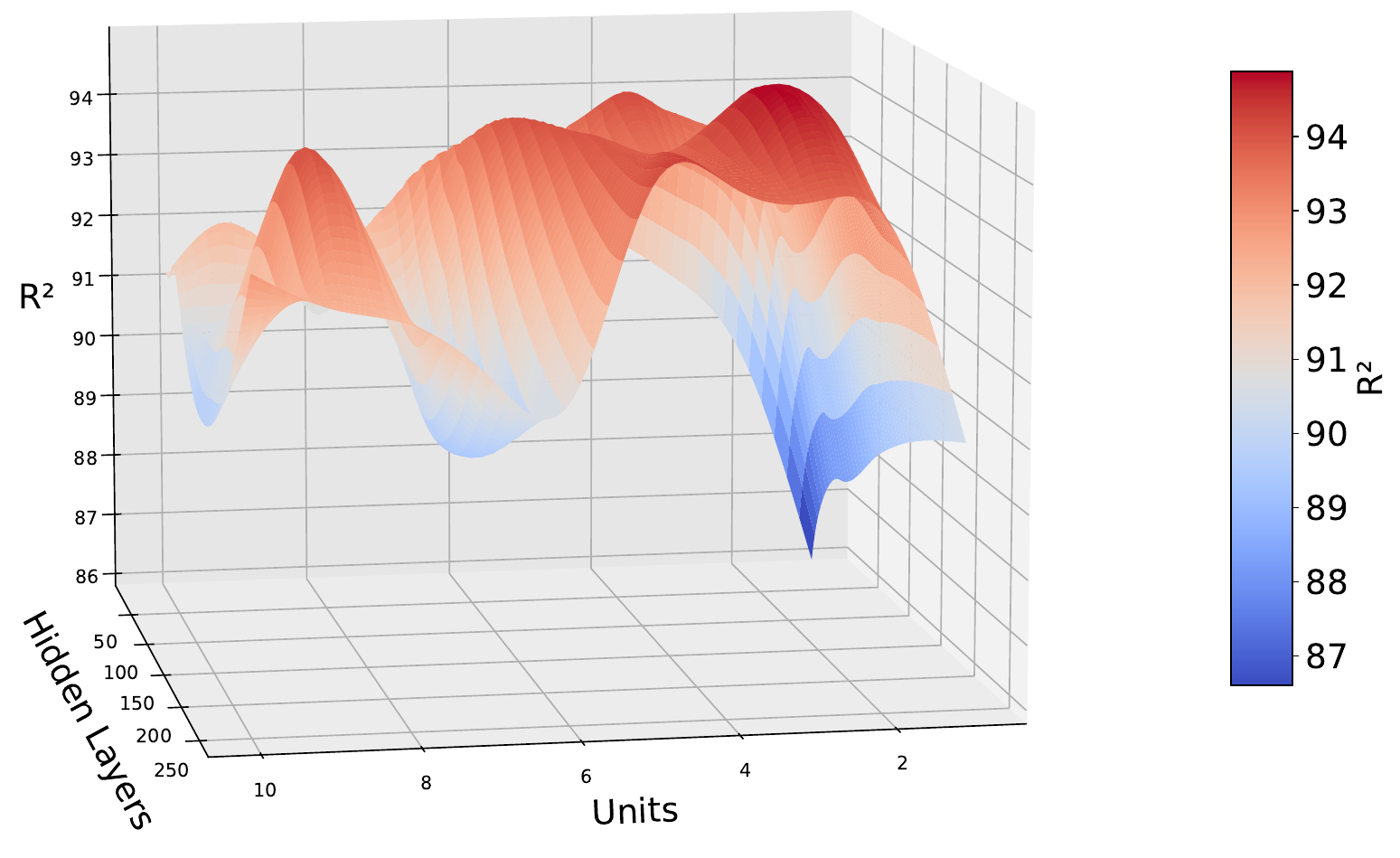}
        \label{fig:subfig-2d}
    }
    \hfill
    \subfloat[]{
        \includegraphics[width=0.485\textwidth, trim={250pt 180pt 160pt 230pt}, clip]{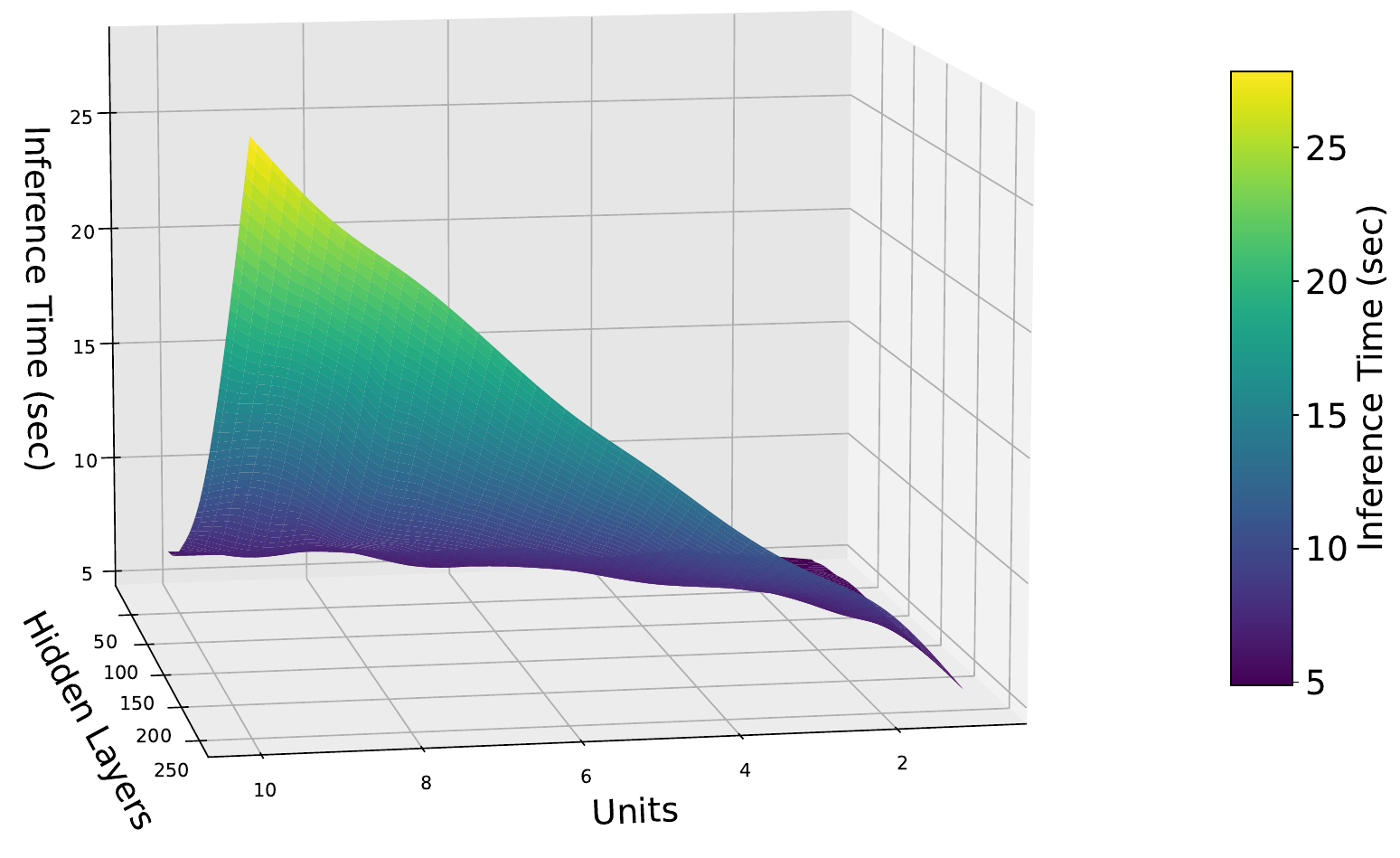}
        \label{fig:subfig-3d}
    }
\caption{Comparative analysis of neural network performance metrics. \textbf{(a)} illustrates how model configurations, varying in units and hidden layers, impact accuracy through the mean R² value. \textbf{(b)} delineates the effect of model complexity on inference time, measured in seconds.}
\label{fig:nn_performance_metrics}
\end{figure*}

Specifically for the ANN, Figure~\ref{fig:nn_performance_metrics} illustrates a comparative analysis of the neural network's performance metrics, demonstrating the best results among the evaluated models. The three-dimensional surface plots capture how various configurations of model units and layers affect both model accuracy, measured by the mean R² value, and the inference time required for predictions. Plot (a) demonstrates that the relationship between the number of units and layers significantly impacts accuracy, indicating an interplay between model depth and performance. Plot (b) shows an increase in inference time with more complex models, emphasizing the need to consider computational efficiency alongside model precision. These visualizations highlight the balance that must be struck in neural network architecture design to optimize both performance and efficiency.

XGBoost achieved moderate accuracy metrics with an MAE of 5.04, MSE of 117.14, RMSE of 10.82, and an R² of 87.31\%. While its error metrics were higher than those of ANN, the differences were not statistically significant according to the Nemenyi test (p-value = 0.2541 for XGBoost vs PR and p-value = 0.2541 for XGBoost vs ANN). XGBoost's performance was acceptable and it may serve as a robust alternative, especially in contexts where ensemble methods are preferred. Its inference time (8.20 seconds) was shorter than PR's but slightly longer than ANN's, with no significant differences detected.

The simplicity of PR resulted in lower predictive performance compared to more advanced models. Its error metrics were significantly higher than those of ANN, as confirmed by the Nemenyi test (p-value = 0.0045 for PR vs ANN). PR lagged behind, with the highest MAE (11.52), MSE (296.26), RMSE (17.21), and the lowest R² value (67.94\%). These results highlight the limitations of PR in capturing complex patterns within the dataset, especially when compared to models like ANN and XGBoost, which are better suited for such complexity.

Note that discrepancies in computational times may be due to differences in library implementations, as PR and ANN were implemented using different libraries, which could impact their computational efficiency.

In summary, the statistical analysis confirms that ANN provides significantly better predictive accuracy compared to PR, without a significant increase in training time. While XGBoost did not show significant differences compared to the other models, it offers a balance between performance and computational efficiency. These findings can guide practitioners in selecting the appropriate model based on the specific requirements of accuracy and computational efficiency, considering both statistical significance and practical performance metrics.

\subsection{Enhanced Weapon Engagement Zone Representation}

\begin{figure}
\centering
\includegraphics[width=0.42\textwidth, trim={1.5cm 1.5cm 1.5cm 1.5cm}, clip]{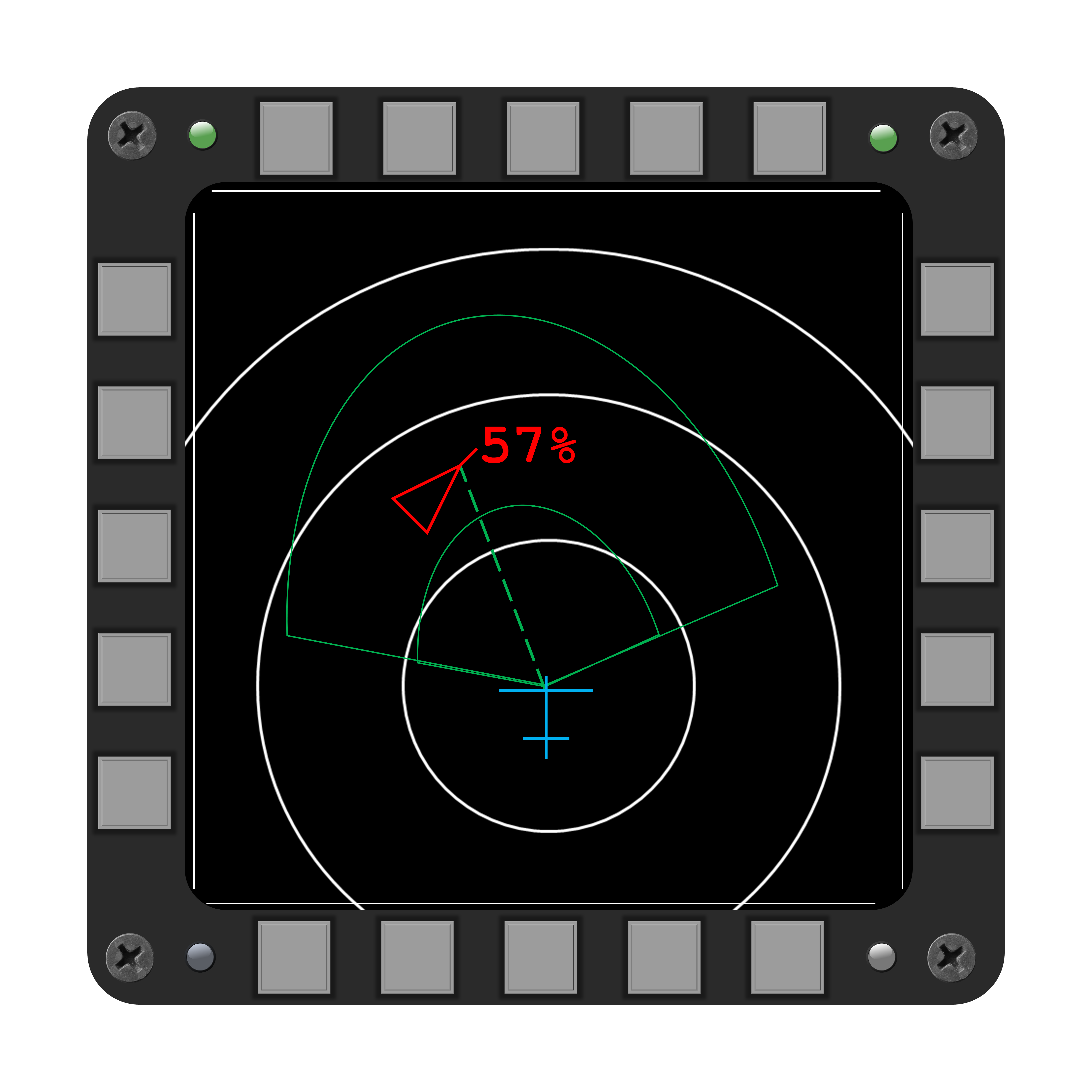}
\caption{Enhanced MFD representation incorporating the integration of $P_{\text{kill}}$ alongside the maximum range and no-escape zones, providing a multifaceted view of the engagement zone to support informed decision-making in missile launch scenarios.} \label{fig: mfd}
\end{figure}

We introduce an enhanced MFD visualization for the WEZ, integrating a probabilistic model that encompasses the traditional maximum range and no-escape zone indicators and a novel probability of kill aspect within the WEZ indications. By applying the proposed probabilistic model, we adjusted the target's off-boresight angles from $-60^{\circ}$ to $+60^{\circ}$. The enhanced WEZ visualization, depicted in Figure~\ref{fig: mfd}, now incorporates $P_{\text{kill}}$, offering a more comprehensive understanding of the engagement zone. Incorporating this metric alongside traditional WEZ metrics may improve a pilot’s situational awareness and decision-making capabilities, particularly in controlled scenarios. However, this potential benefit is heavily dependent on model assumptions and requires further validation for broader applicability. This integration can possibly support pilots in making more informed decisions regarding missile launches against potential targets by providing a clearer assessment of the likelihood of successfully engaging a target under various conditions. Including a probabilistic kill probability metric may aid in optimizing the use of onboard weaponry by evaluating the tactical viability of a missile launch, thus enhancing operational effectiveness and minimizing resource loss.

\section{Conclusion and Future Work}
\label{conclusion}

In conclusion, this work presents PoKER, an innovative probabilistic model for WEZ analysis in BVR air combat scenarios. By extending beyond traditional WEZ metrics to include a probability of kill calculation, PoKER enhances the precision and relevance of missile engagement assessments. This integration may improve the pilot's situational awareness and decision-making capabilities by providing a better understanding of engagement outcomes. The capability of PoKER to offer real-time, accurate estimations of missile effectiveness based on dynamic combat conditions and target behavior represents a notable advancement in air combat strategy and planning. Its application not only can enhance operational efficiency and effectiveness but also contributes to the optimal use of onboard weaponry, thus minimizing resource loss and maximizing engagement success.

For future work, we advocate the utilization of data derived from real training missile launches, whether from virtual simulations or live exercises, to refine the modeling of target behavior. The introduction of human-in-the-loop simulations could significantly enhance the model's practical application by aligning theoretical predictions more closely with real-world decision-making processes. Furthermore, the adoption of advanced machine learning techniques promises to make WEZ modeling more adaptive to changing combat conditions. Enhancements in missile simulation accuracy, alongside investigations into the impact of environmental variables and electronic warfare, could offer deeper insights into engagement strategies. Moreover, developing methods to predict adversary tactics and countermeasures accurately will greatly improve PoKER's strategic utility. Additionally, creating a representation of how to maneuver into regions of higher probability could inform pilots on effective maneuvers before a missile launch.

We also agree that evaluating the dataset size is an important contribution, especially for the scalability of future work. While 10 million simulations were feasible for this study due to available computational resources, we acknowledge that this may not be practical for all scenarios. Therefore, we suggest exploring methods to reduce the dataset size without compromising model fidelity, such as using efficient sampling techniques or selecting a subset of representative points. This could help balance computational costs while maintaining the quality of the analyses.

Furthermore, we plan to conduct empirical studies to validate the overall effectiveness of the proposed model across different scenarios. These studies will provide valuable insights into the model's strengths and limitations in varied operational contexts. While these studies will help assess the model's applicability, incorporating such estimations into real-world systems will require further investigation, which we suggest as a direction for future work as well. The ultimate goal is to demonstrate the utility of PoKER in enhancing situational awareness and decision-making for both autonomous agents and human pilots, particularly in scenarios involving AI-driven adversaries.

By integrating such predictive capabilities, pilots can make more informed decisions, enhancing their engagement strategies. Such efforts can solidify the model's foundational strength and expand its applicability in addressing the complexities of modern air combat. By pushing the boundaries of current WEZ models, this study contributes to academic understanding and offers practical insights for military strategists and defense technology developers.

\begin{sm}
The code and instructions to demonstrate the methodology and calculations used in this study are available at \url{https://github.com/jpadantas/poker}. Due to the classified nature of the data, it is not possible to fully reproduce the results. However, the shared code provides valuable insight into the methods used in this study.
\end{sm}

\begin{acks}
This work has been supported by Finep (Reference nº 2824/20). Takashi Yoneyama and Marcos R. O. A. Maximo are partially funded by CNPq -- National Research Council of Brazil through the grants 304134/2-18-0 and 307525/2022-8, respectively.
\end{acks}

\bibliographystyle{SageV}
\bibliography{references}

\begin{biogs}

\textbf{Joao~P.~A.~Dantas} received his B.Sc. degree in Mechanical-Aeronautical Engineering (2015) and his M.Sc. degree in Electronic and Computer Engineering (2018) from the Aeronautics Institute of Technology (ITA), Brazil. Currently, he is pursuing his Ph.D. degree at ITA and working as a researcher for the Brazilian Air Force at the Institute for Advanced Studies. His research interests include artificial intelligence, machine learning, robotics, and simulation.

\textbf{Andre~N.~Costa} received his B.Sc. degree in Mechanical-Aeronautical Engineering (2014) and his M.Sc. degree in Electronic and Computer Engineering (2019) from the Aeronautics Institute of Technology (ITA), Brazil. Currently, he is a researcher for the Brazilian Air Force at the Institute for Advanced Studies, with research interests in modeling and simulation, artificial intelligence, and machine learning.

\textbf{Diego~Geraldo} received his B.Sc. degree from the Brazilian Air Force Academy (2005) and became a fighter pilot (2006). He earned his M.Sc. degree in Aeronautical and Mechanical Engineering (2012) from the Aeronautics Institute of Technology (ITA), Brazil. Since then, he has been a researcher for the Brazilian Air Force at the Institute for Advanced Studies, specializing in modeling, simulation, and artificial intelligence.

\textbf{Marcos~R.~O.~A.~Maximo} received his B.Sc. degree (Hons.) (summa cum laude) in Computer Engineering (2012), and his M.Sc. (2015) and Ph.D. (2017) degrees in Electronic and Computer Engineering from the Aeronautics Institute of Technology (ITA), Brazil. He is currently a Professor at ITA, being a member of the Autonomous Computational Systems Laboratory (LAB-SCA) with research interests in robotics and artificial intelligence.

\textbf{Takashi~Yoneyama} received his B.Sc. degree in Electronic Engineering (1975) from the Aeronautics Institute of Technology (ITA), Brazil, his M.D. degree in Medicine (1993) from Taubate University, Brazil, and his Ph.D. degree in Electrical Engineering (1983) from Imperial College London, United Kingdom. He is a Professor of Control Theory with the Department of Electronics at ITA. His research is focused on stochastic optimal control theory.
\end{biogs}

\end{document}